\documentclass[runningheads]{llncs}

\usepackage{eccv}

\usepackage{eccvabbrv}

\usepackage{graphicx}
\usepackage{booktabs}
\usepackage{epsfig} 
\usepackage{amsmath}
\usepackage{amssymb}
\usepackage{bm}
\usepackage{amsfonts}
\usepackage{mathrsfs}
\usepackage{pifont}
\usepackage{float}
\usepackage{subcaption}
\usepackage{algorithm}
\usepackage{algorithmic}
\usepackage{bbding}
\usepackage{multicol}
\usepackage{colortbl}
\usepackage{amsmath}
\usepackage{array, makecell}

\DeclareMathOperator*{\argmax}{arg\,max}

\usepackage[accsupp]{axessibility}  

\usepackage{hyperref}
\usepackage[dvipsnames]{xcolor}

\usepackage{orcidlink}

\usepackage{multirow}
\definecolor{lavenderpurple}{RGB}{150,123,182}

\definecolor{yellow}{rgb}{1, 1, 0.7}
\definecolor{orange}{rgb}{1, 0.85, 0.7}
\definecolor{tablered}{rgb}{1, 0.7, 0.7}

\makeatletter
\newcommand{\printfnsymbol}[1]{%
        \textsuperscript{\@fnsymbol{#1}}%
}
\makeatother

\begin{document}

\title{Within the Dynamic Context: Inertia-aware 3D Human Modeling with Pose Sequence}

\titlerunning{Inertia-aware 3D Human Modeling with Pose Sequence}

\author{Yutong Chen\thanks{\footnotesize denotes authors with equal contributions. \textsuperscript{†} denotes co-corresponding authors.}\inst{1}
\and
Yifan Zhan\printfnsymbol{1}\inst{1}\inst{2}
\and
Zhihang Zhong\textsuperscript{†}\inst{1}
\and
Wei Wang\inst{1}
\and
Xiao Sun\textsuperscript{†}\inst{1}
\and\\
Yu Qiao\inst{1}
\and
Yinqiang Zheng\inst{2}
}

\authorrunning{Y.~Chen et al.}

\institute{
Shanghai Artificial Intelligence Laboratory
\and
The University of Tokyo}

\maketitle
\begin{abstract}
    Neural rendering techniques have significantly advanced 3D human body modeling. 
    However, previous approaches overlook dynamics induced by factors such as motion inertia, leading to challenges in scenarios where the pose remains static while the appearance changes, such as abrupt stops after spinning. 
    This limitation arises from conditioning on a single pose, which leads to ambiguity in mapping one pose to multiple appearances.
    
    In this study, we elucidate that variations in human appearance depend not only on the current frame's pose condition but also on past pose states. We introduce Dyco, a novel method that utilizes the delta pose sequence to effectively model temporal appearance variations.
    To mitigate overfitting to the delta pose sequence, we further propose a localized dynamic context encoder to reduce unnecessary inter-body part dependencies.
    To validate the effectiveness of our approach, we collect a novel dataset named I3D-Human, focused on capturing temporal changes in clothing appearance under similar poses. Dyco significantly outperforms baselines on I3D-Human and achieves comparable results on ZJU-MoCap. Furthermore, our inertia-aware 3D human method can unprecedentedly simulate appearance changes caused by inertia at different velocities. The code, data and model are available at our project website at \href{https://ai4sports.opengvlab.com/Dyco}{https://ai4sports.opengvlab.com/Dyco}.
    
    \keywords{3D human \and Pose sequence \and Neural radiance fields \and Inertia}

\end{abstract}
\section{Introduction}
\label{sec:introduction}
Pursuing dynamic, lifelike, and high-fidelity human body reconstruction has been a longstanding endeavor, finding versatile applications in AR/VR, filmmaking, gaming, and motion analysis. 
Previous efforts in dynamic human neural rendering have focused on digitalizing human avatars and modeling human motion~\cite{weng2022humannerf,peng2023implicit,zheng_AvatarRex2023,loper2023smpl,anguelov2005scape,PonsMoll2015_Dyna,qian2023_3dgsavatar,wang2022arah,jiang2023instantavatar,peng2024animatable}. Classical approaches such as SCAPE~\cite{anguelov2005scape} and SMPL~\cite{loper2023smpl} conduct statistical analysis on 3D scans to decompose mesh variation into shape and pose components. However, these methods struggle with lacking geometric details such as clothing depiction. Recent advancements in the neural radiance field~\cite{mildenhall2021nerf} have enabled learning more accurate neural human representations from monocular or multi-view videos, typically involving Linear Blend Skinning (LBS) guided skeletal motion module, MLP-based non-rigid motion module, and canonical radiance fields. These methods achieve pose-dependent human motion modeling by learning human radiance fields in canonical space and warping them to observation space conditioned on pre-extracted poses.

\begin{figure}[t]
    \centering
	\includegraphics[width=1.\columnwidth]{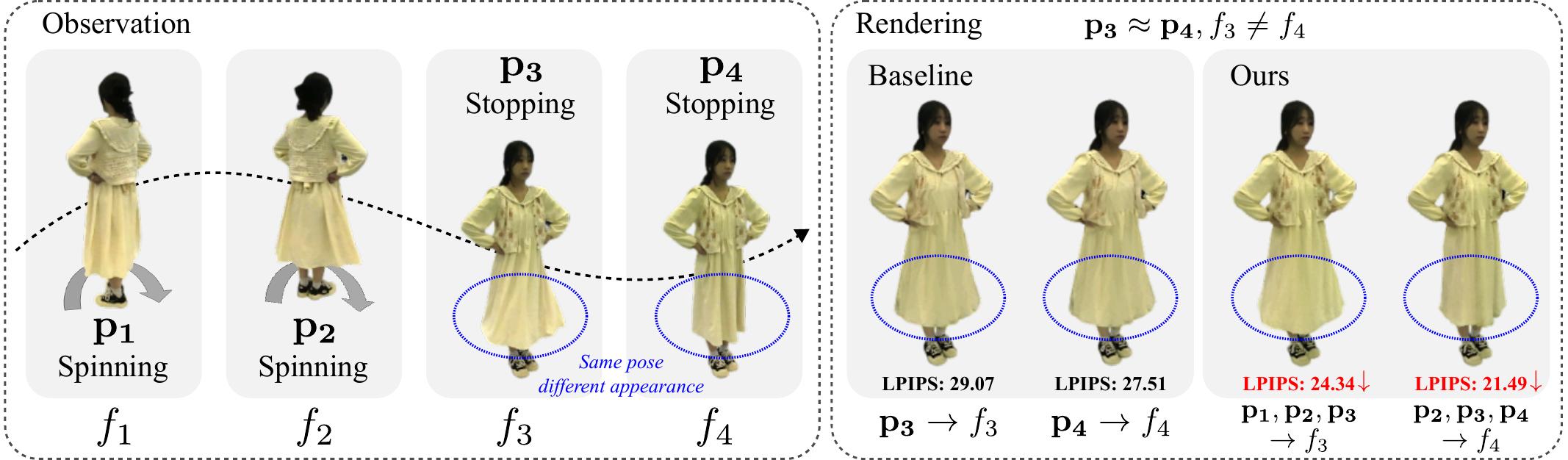}
	\caption{We emphasize that appearance variations not only depend on different static poses, but can also be induced by inertia, such as the graceful hanging down of the dress drape after a sudden stop in motion. Compared with previous methods solely relying on static poses, we encode past pose trajectory with the pose sequence to accurately capture such dynamic effects. This improves both novel-view renderings and generalization to novel poses.}
	\label{fig:teaser}
\end{figure}

\emph{Our question is whether pose-based LBS processes are sufficient for modeling 3D human motion.} 
To answer this question, let us consider the scenario captured from a static camera view in \cref{fig:teaser}, wherein a girl spins ($\bm{p}_1\to \bm{p}_3$) and then stops ($\bm{p}_3 \to \bm{p}_4$). As she stops, her pose stays mostly the same, but the drape of her dress noticeably falls.
Therefore, solely relying on the current single pose is insufficient to model variations in non-rigid human motion and appearance, as it ignores dynamic motion contexts. This limitation causes previous methods to struggle with mapping identical pose inputs to different appearance, resulting in artifacts in the scenario described above.

In this work, we propose a neural avatar model that accounts for dynamic context, dubbed Dyco, which significantly advances dynamic human motion modeling by resolving the ambiguity that arises from relying solely on static poses. Drawing inspiration from dynamic blend shapes in DMPL~\cite{loper2023smpl}, we encode pose sequences as condition in addition to the current static pose. We compute the changes, or deltas, in pose parameters over the pose trajectory and input the delta sequence into the non-rigid deformation field and canonical volume. Since the increased input dimension poses a risk of overfitting, we design a localized dynamic context encoder to reduce input complexity without sacrificing information. The dynamic context encoder eliminates spurious dependencies between different body parts and efficiently aggregates spatial and temporal information.

In addition to the methodological neglect of dynamic contexts, current datasets fail to accurately capture the associated issues. 
existing datasets mitigate the issue of appearance variations caused by dynamic contexts by reducing the magnitude of motion (DNA-Rendering Dataset~\cite{cheng2023dna}) or wearing tight-fitting garments (ZJU-Mocap Dataset~\cite{peng2021_neuralbody}, ENeRF-Outdoor Dataset~\cite{lin2022efficient} and NHR Dataset~\cite{wu2020multi}). 
However, real-world human body modeling cannot avoid dealing with large motions and loose clothing.  
To fully demonstrate the advantages of our inertia-aware 3D human modeling method, we have collected a new dataset of human body data, named I3D-Human. 
I3D-Human emphasizes significant appearance variations resulting from dynamic contexts, such as skirt swaying and hem dropping, with the aim of facilitating relevant future research.
In summary, our contributions are as follows:
\begin{itemize}

  \item [1)]
    Dyco, a novel human motion modeling method that utilizes pose sequences, mitigating the ambiguity in appearance learning conditioned solely on poses, thus enhancing the generalization of dynamic contexts to novel poses;
  \item [2)]
    a localized dynamic context encoder to alleviate the risk of overfitting associated with the increased representational capacity, thereby contributing to improved model performance;
 \item [3)]
    I3D-Human Dataset, a self-collected human dataset capturing richer variations of appearance from dynamic contexts. We achieve best results on our I3D-Human Dataset and demonstrate comparable performance on ZJU-Mocap, compared to state-of-the-art human body modeling methods.
\end{itemize}
\section{Related Work}
\label{sec:related_work}

\subsection{Explicit Mesh Representation for Human Avatars}
The early success of digital avatars took place in a class of body models~\cite{Hirshberg_SCAPE,loper2023smpl, Freifeld_LieBodies, PonsMoll2015_Dyna} that represent the body as a mesh and perform statistical analysis on the database of 3D scans to learn shape and pose components in the mesh variation. Mostly well-known, 
a family of SMPL(-X)-based human body models~\cite{loper2023smpl} has been widely used over the past decade for its flexibility, realism, and ease of use. The key improvement of SMPL over previous models lies in its effective learning of \emph{pose blend shapes} to correct skinning artifacts and pose-dependent deformations. This is achieved by formulating the pose blend shapes as a linear function of the elements of the part rotation matrices.
Since then, plenty of work has contributed to estimating the SMPL parameters of subjects from 2D inputs~\cite{dong2021fast,easymocap,shuai2022multinb}.  To capture the momentarily lagging-behind motion of soft body tissues in dynamic human motion, DMPL\cite{loper2023smpl} and Dyna\cite{PonsMoll2015_Dyna} parameterized velocities and acceleration of the body and limbs into their linear models, which brings their animation closer to life. However, these meshed-based body models can only represent naked or tightly-clothed bodies. Though some work~\cite{guan2012drape,patel20_tailornet} extends body mesh models to modeling clothed humans, their mesh-based models of limited resolution make the renderings far from visually realistic.

\subsection{Implicit Neural Representation for Human Avatars}
In pursuit of photorealism, advanced human avatars exploit neural implicit representation~\cite{Yu2021_ANeRF,Liu2021_NeuralActor,li2022tava,zheng_AvatarRex2023}. For instance, a number of methods using neural radiance field (NeRF~\cite{mildenhall2021nerf}), ARAH~\cite{wang2022arah} using neural signed distance function (NeuS~\cite{wang2021neus}), InstantAvatar~\cite{jiang2023instantavatar} and FastNeuralBody using InstantNGP~\cite{muller2022instant}, and more recent models take advantage of the efficiency of gaussian splatting~\cite{kerbl20233d}.  Despite variations in their choice of neural representation, these methods typically learn a canonical field to represent body shape and appearance in a rest pose, along with a deformation module to compute correspondence between the rest pose and target pose. 

However, while neural 3D representations have demonstrated impressive results in novel-view rendering of static scenes and even some quasi-static dynamic scenes, we find that the rendering results of neural implicit human avatars are not satisfactory enough, exhibiting blurring appearance and failing to manifest realistic dynamic motion, even in the training images.  One important reason for the blurriness is that most of the state-of-arts only take the SMPL parameters of the target single pose as input to their neural representation, which will bring dynamic context-induced ambiguity that may influence model fitting. 

\subsection{Modeling Dynamic Context in Human Avatars}
The impact of dynamic context on human appearance was discussed in some previous literature. First, there are physical-based approaches which exploited the material properties of human soft tissues and simulated their dynamic simulation using intricate physical models.
For example, Larboulette \etal~\cite{Larboulette_DynamicSkinning} design a mechanical model of a regular limb to simulate the motion of flesh under different accelerations of the bone with the finite element method (FEM). Rohmer \etal~\cite{Rohmer2021VelocitySF} underscores the importance of secondary animation for liveliness and presents velocity skinning on top of standard LBS to create dynamic effects such as squash and stretch.  However, physical-based approach requires manual design for different materials, which is infeasible to apply to full-body animation. Instead of modeling the underlying physics, another line of work attempts to learn the approximation of dynamics from 4D scans in a data-driven manner. DMPL~\cite{loper2023smpl} and Dyna~\cite{PonsMoll2015_Dyna} learn to predict soft-tissue deformation from the changing of the body pose. However they can only model naked human bodies. DRAPE~\cite{guan2012drape} incorporates motion history into clothing deformation prediction to animate the draping and flowing effect of clothing. However, it requires the clothing template mesh and its proficiency is highly limited to the types of clothing it is trained on. Also, these methods are confined to mesh-based representation and deformation and thus far from high-resolution photo-realistic renderings. 

On the other hand, current implicit neural avatars are promising for photo-realisim. However, none of them manage to represent the effect of dynamic context and motion inertia on appearance variation. Our work is the first to make the implicit avatar inertia-aware. One reason for the dynamic effect being overlooked is that current datasets, \eg ZJU-MoCap~\cite{peng2021_neuralbody}, PeopleSnapshot~\cite{Alldieck2018_PeopleSnapShot}, only focus on tightly-clothed figures with slow and gradual motion. To study the dynamic motion effects that commonly exist in the real scenario, we curate a new dataset of loosely-clothed humans with free and abrupt motion.

\section{Preliminaries}
\label{sec:preliminaries}

\subsection{Human NeRF Representation}

A HumanNeRF~\cite{weng2022humannerf} is defined to determine the density $\sigma$ and color $\bm{c}$ of any given 3D location $\bm{x}$, conditioned on a specific pose configuration $\bm{p} = \{(R_i, \bm{t}_i)|i \in K\}$ of a person: $F_o(\bm{x}, \bm{p}) \xrightarrow{}(\sigma, \bm{c})$.
Typically, the problem is solved in a canonical space, where a standard static Neural Radiance Field (NeRF)~\cite{mildenhall2021nerf}
\begin{equation}
    F_c(\bm{x}_c) \xrightarrow{\Theta_{\textrm{NeRF}}}(\sigma, \bm{c}),
\end{equation}
and a point mapping function $T(\bm{x}, \bm{p}) \xrightarrow{}\bm{x}_c$ from observed space back to canonical space are learned.
The $\bm{x}_c$ can be split into a rigid and a non-rigid part, \ie, $\bm{x}_r$ and $\Delta \bm{x}$, corresponding to the rigid transformation function $T_\textrm{R}$ and non-rigid transformation function $T_\textrm{NR}$, represented as
\begin{align}
    \bm{x}_c &= \bm{x}_r + \Delta \bm{x},\\
    &= T_\textrm{R}(\bm{x}, \bm{p}) + T_\textrm{NR}(\bm{x}_r, \bm{p}).
\end{align}

The rigid transformation function $T_\textrm{R}$ is formulated as an inverse linear blend skinning process~\cite{lewis2023pose}, a weighted sum of the rigid transformations defined following
\begin{equation}
    T_\textrm{R} = \sum_{i=1}^{K}\omega_o^i(\bm{x})(R_i\bm{x} + \bm{t}_i),
\end{equation}
where $\omega_o^i(\bm{x})$ is the corresponding blending weight in observed space and estimated using a set of learnable weights $\omega_c^i(\bm{x})$ defined in the canonical space as

\begin{equation}
    \omega_o^i(\bm{x}) = \dfrac{\omega_c^i(\bm{x})(R_i\bm{x} + \bm{t}_i)}{\sum_{i=1}^{K}\omega_c^i(\bm{x})(R_i\bm{x} + \bm{t}_i)}.
\end{equation}
Using a convolution network that takes a random latent code $\bm{z}$ as input, a single volume grid $W_c(\bm{x})$ with $K+1$ channels are learned to store all the blending weights $W_c(\bm{x}) = \{\omega_c^i(\bm{x})|i \in K\}$: $\textrm{CNN}_{\Theta_\omega}(\bm{z}) \xrightarrow{}W_c(\bm{x})$.

The non-rigid transformation $T_\textrm{NR}$ predicts an extra position offset $\Delta\bm{x}$ caused by different pose configurations by optimizing parameters $\Theta_\textrm{NR}$,
\begin{equation}
    T_\textrm{NR}(\bm{x}_r, \bm{p}) \xrightarrow{\Theta_\textrm{NR}} \Delta \bm{x}.
\end{equation}

\subsection{NeRF and Volume Rendering Revisited}

NeRF~\cite{mildenhall2021nerf} comprises three primary stages: sampling, volume mapping, and rendering. In the sampling stage, points $\bm{x}\in\mathbb{R}^3$ are sampled along rays computed from the camera's position. Subsequently, in the volume mapping stage, each 3D point $\bm{x}$, along with its viewing direction $\bm{d}\in\mathbb{R}^3$, is queried to determine the volume density $\sigma$ and color $\bm{c}=(r,g,b)$ of $\bm{x}$. Finally, in the rendering process, the color of each ray is determined using volume rendering techniques~\cite{kajiya1984ray}. The expected color of the ray $\bm{r}(t)=\bm{o}+t\bm{d}$ is then computed as follows 
\begin{equation}
  C(\bm{r})=\int_{t_n}^{t_f}\ T(t)\sigma(\bm{r}(t))\bm{c}(\bm{r}(t),\bm{d})dt\,,
  \label{eq:Rendering equation}
\end{equation}
where $t_n$ and $t_f$ are the near and far bounds, respectively, and 
\begin{equation}
    T(t)=\exp\left(-\int_{t_n}^{t}\ \sigma(\bm{r}(s))ds\right)\,.
\end{equation}
The ground truth color $C_\textrm{gt}(\bm{r})$ of each ray serves as supervision for training NeRF, and we define the loss function as follows
\begin{equation}
  \mathcal{L}_{\textrm{MSE}
  }=\sum_{\bm{r}\in\mathcal{R}}\ \left\|C(\bm{r})-C_\textrm{gt}(\bm{r})\right\|_2^2\,,
  \label{eq:loss function}
\end{equation}
where $\mathcal{R}$ is the ray batch. Besides RGB-based loss, previous work~\cite{weng2022humannerf,yu2023monohuman,qian2023_3dgsavatar} shows that utilizing an LPIPS loss~\cite{zhang2018_lpips} $\mathcal{L}_{\textrm{LPIPS}}$ as an auxiliary supervision can greatly enhance the visual details of the renderings.

\section{Method}
\label{sec:method}
\subsection{Ambiguities under Identical Pose Conditions}
\label{sec:Ambiguities}
Relying solely on rigid transformations, $T_{R}$, for modeling details of human motion is not sufficient.
In many human NeRF works~\cite{li2022tava,weng2022humannerf}, achieving comprehensive motion modeling requires incorporating non-rigid deformations $T_{NR}(\bm{x}_r, \bm{p})$ conditioned on human poses $\bm{p}$. Some studies~\cite{noguchi2021neural} introduce additional pose conditions for training the canonical volume to capture variations in per-frame appearance.
Nevertheless, our analysis suggests that relying exclusively on pose conditions for non-rigid deformation and canonical volume learning still imposes inherent ambiguities. This is exemplified by varying clothing states of a person under identical poses. In instances where an individual wearing loose clothing abruptly halts after rotation, there is a brief period during which the overall pose remains static, yet inertia induces substantial swinging motion at the specific location $\bm{x}$ on the clothing. Consequently, the non-rigid deformation target $\Delta\bm{x}$ for point $\bm{x}$ becomes ambiguous due to this swinging motion, given their shared condition under identical poses. Another type of ambiguity arises from the variations in appearance under identical pose conditions, when two observation points corresponding in canonical space exhibit different local normal characteristics. Even under invariant lighting conditions, differences in local normals can impact shading outcomes, consequently affecting the final appearance. 

We emphasize that these two ambiguities are prevalent in everyday scenarios caused by dynamic contexts such as inertia, despite being overlooked by previous methods and datasets focusing on meticulously designed wearing patterns. To alleviate these ambiguities, we introduce pose-sequence information in the following sections to capture the intrinsic details arising from inertia.

\begin{figure}[ht]
    \centering
	\includegraphics[width=1\columnwidth]{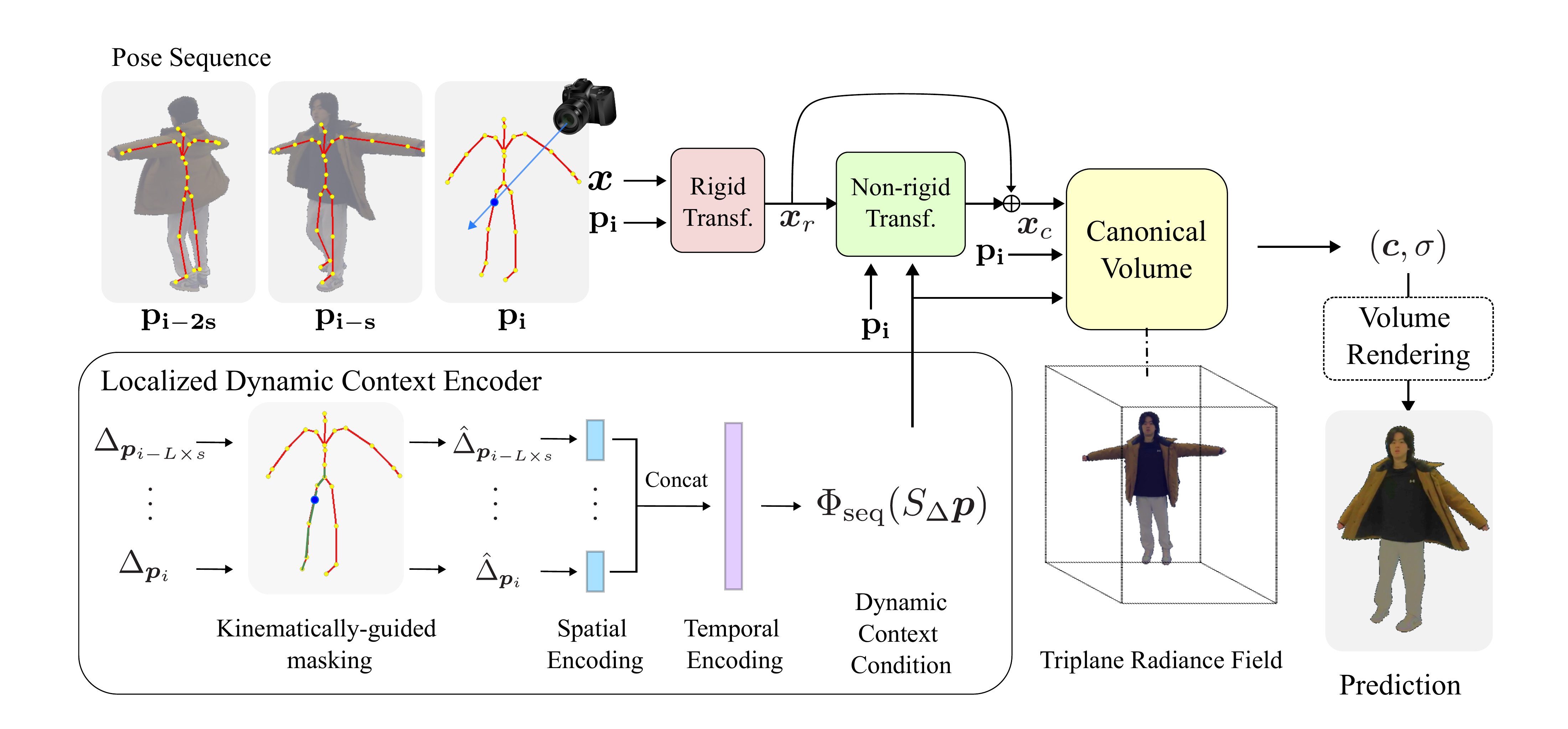}
	\caption{The overall pipeline of our method. The rigid transformation and non-rigid transformation module deform the coordinate in the pose space into the canonical space, which is then fed into the triplane volume to obtain the color and density in the canonical space. To capture the variation under similar poses within different dynamic contexts, we adopt a localized dynamic context encoder to embed pose sequences as additional conditional inputs into the transformation module and canonical volume. }
	\label{fig:method}
\end{figure}

\subsection{Modeling Pose-Sequence Dependent Human Motion}
\cref{fig:method} shows our pipeline. Our core insight is that the inertia properties of the human body cannot be exclusively inferred from the current frame but are encoded within the preceding pose sequence. We introduce two parameters to define such a pose sequence: sequence length $L$ and step $s$. Formally, a pose-sequence $S_{\bm{p}}$ with a length $L$, traced back from the current frame at step $s$, is defined as
\begin{equation}
    S_{\bm{p}} = \{\bm{p}_{i-(L-1)s},\ldots,\bm{p}_{i-s},\bm{p}_i\},
\end{equation}
where the sequence length $L$ is the number of frames in the pose-sequence and $s$ is the level of granularity. Both parameters jointly determine the pose sequence's window length and motion details of the pose sequence. However, this pose-sequence design is susceptible to overfitting when used as network input, particularly due to its complexity with increasing sequence length $L$.

We reassess the inertia-related properties, \eg velocity and acceleration, of the human body and find that they are solely associated with \textbf{changes} in pose, independent of absolute pose. Therefore, we further define a delta-pose sequence with length $L_{d}$, sequence step $s$, and delta step $s_{d}$ as
\begin{align}
    S_{\Delta{\bm{p}}} = \{\Delta{\bm{p}_{i-(L_{d}-1)s}},&\Delta{\bm{p}_i},\ldots,\Delta{\bm{p}_{i-s}}, \Delta{\bm{p}_i}\}\,,\\
    \Delta{\bm{p}_i} &= \delta(\bm{p}_i, \bm{p}_{i-s_{d}}),
\end{align}
where $\delta$ is a $3K+3$ dimensional vector including the per-joint rotation difference represented in axis-angle form between two poses and the global translation difference. The proposed $S_{\Delta{\bm{p}}}$ reduces input complexity by only encoding pose residuals. The sequence vector is then embedded by a dynamic context encoder $\Phi_{\textrm{seq}}$ which will be explained in~\cref{sec:dynamic_encoder}.

As $S_{\Delta{\bm{p}}}$ only encompasses temporally local pose difference, we supplement the context condition with the current pose $\bm{p}_i$. Altogether, we utilize both the current pose condition and the context sequence to guide the learning of the non-rigid deformation and the canonical volume, effectively resolving the appearance ambiguities caused by inertia. For non-rigid deformation, we have

\begin{equation}
    T_{NR}(\bm{x}_r,\bm{p}_i,\Phi_{\textrm{seq}}(S_{\Delta{\bm{p}}})) \xrightarrow{\Theta_{\textrm{NR}}} \Delta \bm{x}\,,
\end{equation}

\noindent and for canonical radiance field, we formulate 

\begin{equation}
\label{ref:eq_Fc-with-condition}
    F_c(\bm{x}_c, \bm{p}_i,\Phi_{\textrm{seq}}(S_{\Delta{\bm{p}}})) \xrightarrow{\Theta_{\textrm{NeRF}}}(\sigma, \bm{c})\,.   
\end{equation}

As MLP-based canonical representations suffer from slow convergence and require a deep deformation network to model human motion, we employ a triplane-based low-rank model~\cite{Chan2022_Eg3D} to represent 3D canonical space. This representation allows for rapid convergence while significantly reducing the depth of non-rigid deformation network ($2$ layers in our implementation), making it possible to achieve fast and high-quality canonical space reconstruction.

\subsection{Localized Dynamic Context Encoder}
\label{sec:dynamic_encoder}

Our designed pose-sequence-related condition enables the model to express variation for the same pose within different inertia states. However, it may also lead the model to learn spurious correlations between the high-dimensional input and the output. To address this issue, we propose a localized dynamic context encoder $\Phi_{\textrm{seq}}$ to embed the input context sequence.

\smallskip \noindent \textbf{Kinematically-guided Spatial Dependency:} For a given context sequence $S_{\Delta{\bm{p}}}$ and an input coordinate $\bm{x}$, we begin by processing each $\Delta{\bm{p}}\in \mathbb{R}^{3\times K+3}$ independently. We leverage a reasonable physical assumption: the impact of a rotating joint is mainly restricted to the kinematic chains to which it belongs, so the motion and appearance of a point $\bm{x}$ should mainly depend on the joints within the same kinematic chains. With an input coordinate $\bm{x}$ and its associated blending weights $W_c(\bm{x})\in \mathbb{R}^{K}$, we first identify its nearest joint $k=\argmax W_c(\bm{x})$ to determine the kinematic chains $\bm{x}$ lies on. Then we gather all joints from the related kinematic chains and apply a binary mask to $\Delta_{\bm{p}}$ to filter out the joints outside the kinematic chains. We denote the masked conditions as $\hat{\Delta}{\bm{p}}$ and the masked sequence as $S_{\hat{\Delta}\bm{p}}$. 

\smallskip \noindent \textbf{Spatial-Temporal Encoding}:
For $S_{\hat{\Delta}\bm{p}}$, we initially employ an MLP to reduce the spatial dimension from \( 3 \times K + 3 \) to 16, resulting in a vector sequence size of \( L \times 16 \). To integrate temporal information, we flatten the sequence and apply a second MLP to project it to a vector dimension of 32. The resulting 32-dimensional vector is concatenated with the positional embedding of the coordinate and serves as input to the subsequent radiance field.

\subsection{I3D-Human Dataset and Dynamic Human Evaluation}
\label{sec: I3D_Dataset}
Human motion in lightweight wearings and at variable speeds is ubiquitous in daily lives. However, current multi-view avatar benchmarks such as ZJU-MoCap~\cite{peng2021_neuralbody}, PeopleSnapshot~\cite{Alldieck2018_PeopleSnapShot} and Human3.6M~\cite{Ionescu2014_Human3.6M}, are collected under controlled speeds and with tight-fitting garments, turning off inertia-induced dynamic effects. This limitation inhibits us from testing the model's ability to capture real-life dynamic effects. To address this gap and advance human avatars toward greater photorealism, we curated a new dataset named the Inertia-aware 3D Human (I3D-Human) dataset. Additionally, we design a new motion-based metric for evaluation purposes.

\smallskip \noindent \textbf{The I3D-Human Dataset: }The dataset focuses on capturing variations in clothing appearance under approximately identical poses. Compared with existing benchmarks, we outfit the subjects in loose clothing such as dresses and light jackets and encourage movements involving acceleration or deceleration, such as sudden stops after spinning, swaying, and flapping sleeves. Our capturing equipment consists of 10 DJI Osmo Action cameras, shooting at a frame rate of 100fps while synchronized with an audio signal. The final processed dataset records 10k frames of sequence from 6 subjects in total. More details about the I3D-Human Dataset can be found in our supplemental materials. 

\smallskip \noindent \textbf{Evaluating Dynamic Motion Effects:} Traditional per-frame metrics such as PSNR measure static visual fidelity. Here we introduce a motion-based metric named dynamic motion error (DME) to assess the faithful portrayal of dynamic effects. While we lack ground-truth 3D motion trajectories, we can utilize optical flow to assess motion in pixel space. Formally, we denote the rendered image at the $i$-th time-step as $\bm{C}^{i}$ and its ground truth as $\bm{C}_{\textrm{gt}}^{i}$. We utilize the off-the-shelf optical flow estimator RAFT~\cite{Teed2020_RAFT} to calculate the pixel translations between two steps in predictions and ground truths as follows:
\begin{align} \label{eq1}
\bm{f}_\textrm{gt}^i &= \textrm{RAFT}(\bm{C}_\textrm{gt}^{i-1}, \bm{C}_\textrm{gt}^{i}), \ \\
\bm{f}^i &= \textrm{RAFT}(\bm{C}^{i-1}, \bm{C}^{i}). \
\end{align}
Then we compute the end-to-end point error (EPE) between $\bm{f}_\textrm{gt}^i$ and $\bm{f}^i$. The final DME for a whole test sequence is the average EPE across all time-steps:
\begin{equation}
 \textrm{DME} = \frac{1}{T}\sum_{i=2}^{T}\left\|\bm{f}_\textrm{gt}^i-\bm{f}^i\right\|.
\end{equation}
\section{Experiments}
\label{sec:experiments}
\subsection{Datasets}
Given our central objective of imbuing neural implicit avatars with inertia-aware dynamic effects, we mainly examine and compare the methods on the I3D-Human dataset. We conduct experiments on 4 sequences from the dataset and split each sequence into around 600 training frames and 400 test frames. We take 4 cameras for training and the rest for testing. As it remains crucial to maintain performance in the scenario of tightly-clothed subjects in slow and controlled motion, we also compare our method with others on the mostly used ZJU-MoCap~\cite{peng2021neural} benchmark. ZJU-MoCap comprises 9 multi-view RGB sequences. Following the training setup in previous work~\cite{peng2021neural,peng2021animatable,peng2024animatable}, we use 4 cameras for training and the rest 19 cameras for testing. Each sequence is divided into two consecutive clips, one for training and the other for testing unseen poses. 
\setlength\tabcolsep{.7em}
\begin{table}[t!]
\centering
\caption{\textbf{Comparison with state-of-the-arts on I3D-Human}: The scores are averaged over the four sequences. Dyco outperforms all baselines across all metrics, particularly excelling in the Dynamic Motion Error (DME). $\textrm{LPIPS}^*=\textrm{LPIPS}\times10^3$}
\resizebox{\textwidth}{!}{
\begin{tabular}{lcccccccc}
\toprule
 \multirow{2}{*}{Method} & \multicolumn{4}{c}{Novel View} & \multicolumn{4}{c}{Novel Pose} \\
 \cmidrule(lr){2-5} \cmidrule(lr){6-9}
    & PSNR$\uparrow$ & SSIM$\uparrow$ & $\textrm{LPIPS}^*\downarrow$ & DME$\downarrow$ & PSNR$\uparrow$ & SSIM$\uparrow$ & $\textrm{LPIPS}^*\downarrow$ & DME$\downarrow$ \\
\midrule    
NeuralBody~\cite{peng2023implicit}     & 30.33 & 0.9681 & 60.89 & 5.08 & 28.80 & 0.9604 & 67.59 & 4.63 \\
AniNeRF~\cite{peng2024animatable}     & 29.29 & 0.9662 & 61.95 & 5.73 & 28.48 & 0.9628 & 64.85 & 4.39 \\
AniSDF~\cite{peng2024animatable}    & 29.20 & 0.9670 & 58.94 & 5.56 & 28.34 & 0.9632 & 62.18 & 4.48 \\
HumanNeRF~\cite{weng2022humannerf}& 29.53 & 0.9678 & 42.04 & 5.25  &  28.78  & 0.9644 & 45.70 & 4.39 \\
3DGS-Avatar~\cite{qian2023_3dgsavatar}&30.62 & 0.9712 & 39.74 & 4.93 & 29.21 & 0.9658 & 44.61 & 4.26 \\
Dyco (Ours) & \textbf{31.22} & \textbf{0.9738} & \textbf{34.54} & \textbf{4.52}  &  \textbf{30.12}  & \textbf{0.9691} & \textbf{39.55} & \textbf{3.98} \\
\bottomrule
\end{tabular}
}
\label{table:sota_I3DH}
\end{table}

\setlength\tabcolsep{.7em}
\begin{table}[t!]
\centering
\caption{ \textbf{Comparison with state-of-the-arts on ZJU-MoCap}: The scores are averaged over the 9 sequences. Dyco achieves comparable results with state-of-the-arts, preserving the ability of modeling tightly-clothed humans in slow and gradual motion.  $\textrm{LPIPS}^*=\textrm{LPIPS}\times10^3$}


\resizebox{\textwidth}{!}{

\begin{tabular}{lcccccccc}
\toprule
 \multirow{2}{*}{Method} & \multicolumn{4}{c}{Novel View} & \multicolumn{4}{c}{Novel Pose} \\
 \cmidrule(lr){2-5} \cmidrule(lr){6-9}
    & PSNR$\uparrow$ & SSIM$\uparrow$ & $\textrm{LPIPS}^*\downarrow$ & DME$\downarrow$ & PSNR$\uparrow$ & SSIM$\uparrow$ & $\textrm{LPIPS}^*\downarrow$ & DME$\downarrow$ \\
    
\midrule   
NeuralBody~\cite{peng2023implicit}&\textbf{33.59} & \textbf{0.9804 } & 36.03 &\textbf{5.02}& 29.68 & 0.9656 & 49.70 &7.22 \\
AniNeRF~\cite{peng2024animatable}&32.28 & 0.9750 & 47.53 & 5.44 & 29.86 & 0.9667 & 52.61 & 6.91 \\
AniSDF~\cite{peng2024animatable}&32.16 & 0.9767 & 40.63 & 5.60 & 29.58 & 0.9677 & 45.63 & 6.96 \\
HumanNeRF~\cite{weng2022humannerf}& 32.13 & 0.9783 & 22.30 &5.78&  29.62 & 0.9678 & 31.11 &7.43\\
3DGS-Avatar~\cite{qian2023_3dgsavatar} &32.81 & 0.9791 & 23.76 &5.47& \textbf{30.01} & \textbf{0.9699} &\textbf{30.20} &\textbf{6.98}\\
Dyco (Ours) & 32.80 & 0.9800 &\textbf{20.32} &5.20&  29.71 & 0.9681 & 31.22&7.22 \\
                     \bottomrule
\end{tabular}
}
\label{table:sota_zju}
\end{table}

\subsection{Evaluation}
We compare the methods on two test splits, namely novel view and novel pose. Novel-view test set evaluates the observed poses in the training frames from novel camera views, assessing whether the model can learn a physically realistic 3D human under the training poses from the 2D RGB observations. Novel-pose test set renders unobserved poses from novel camera views, examining whether the learned 3D avatars can be plausibly animated under unseen poses. For evaluation metrics, we adopt traditional per-frame metrics including PSNR, SSIM, and LPIPS~\cite{zhang2018_lpips}. As explained in~\cref{sec: I3D_Dataset}, we also propose a novel motion-based metric termed DME to measure the accuracy of motion dynamics rendering.

\subsection{Comparison with State of the Arts}
We compare with state-of-the-art counterparts, including Neural Body~\cite{peng2023implicit}, AniNeRF~\cite{peng2024animatable}, AniSDF~\cite{peng2024animatable}, HumanNeRF~\cite{weng2022humannerf} and 3DGS-Avatar~\cite{qian2023_3dgsavatar}. The last two models are initially proposed for the monocular training setting so we adapt them to our multi-view setting. Note that the concurrent 3DGS-Avatar harnesses the advanced capabilities of Gaussian splatting representation~\cite{kerbl2023_3dgs}.

\begin{figure}[t!]
  \centering
    \includegraphics[width=\textwidth]{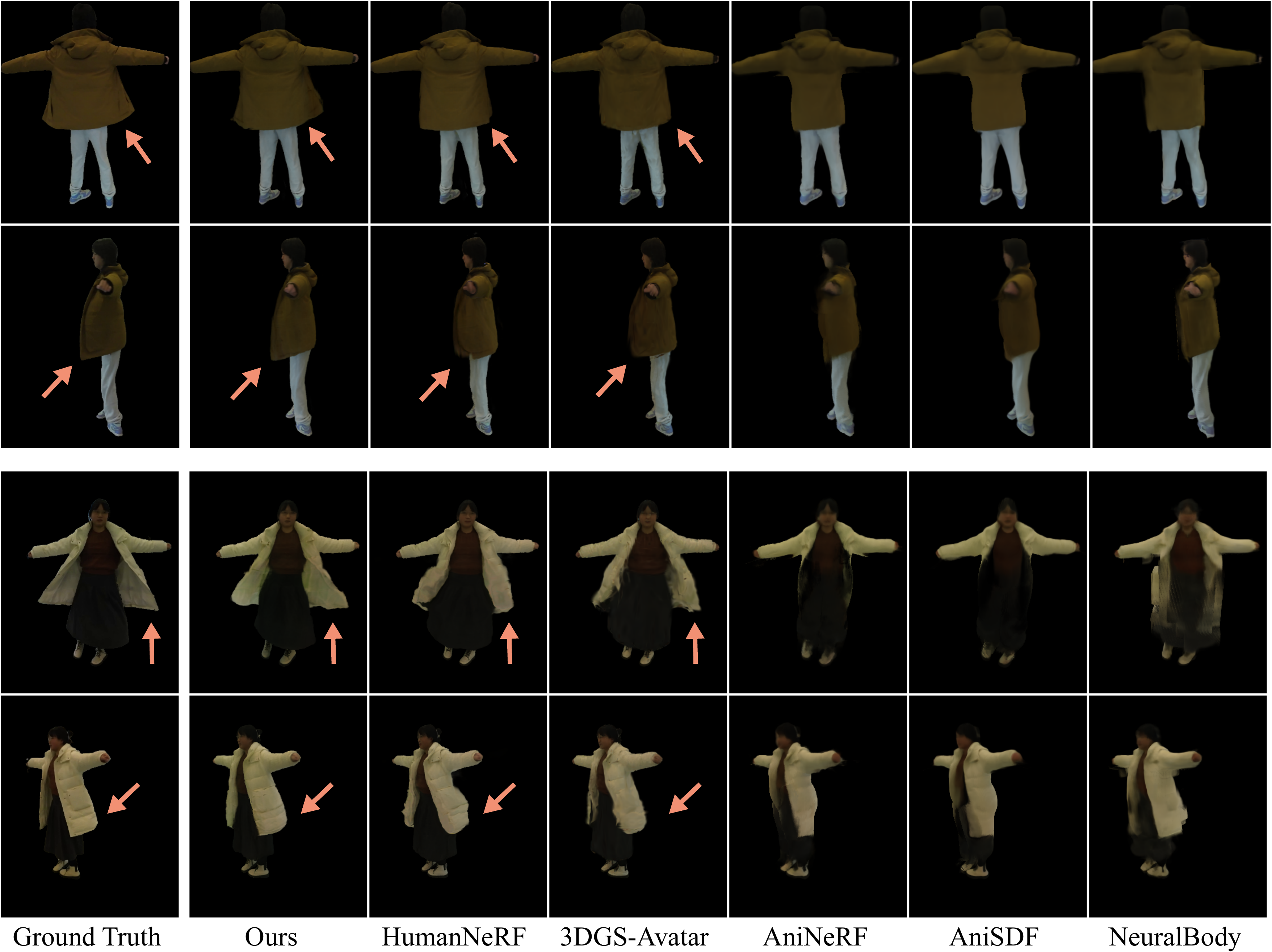}
    \caption{\textbf{Qualitative comparison on the novel view of I3D-Human dataset.} Our method can render distinct appearance and deformation across various motion contexts. In contrast, HumanNeRF~\cite{weng2022humannerf} and 3DGS-Avatar~\cite{qian2023_3dgsavatar} struggle to capture the precise details. The other three baselines~\cite{peng2024animatable,peng2023implicit} exhibit noticeable artifacts.}
    \vspace{-5mm}
    \label{fig:i3d_qualitative_cmp_novelview}
\end{figure}

\begin{figure}[t!]
  \centering
    \includegraphics[width=\textwidth]{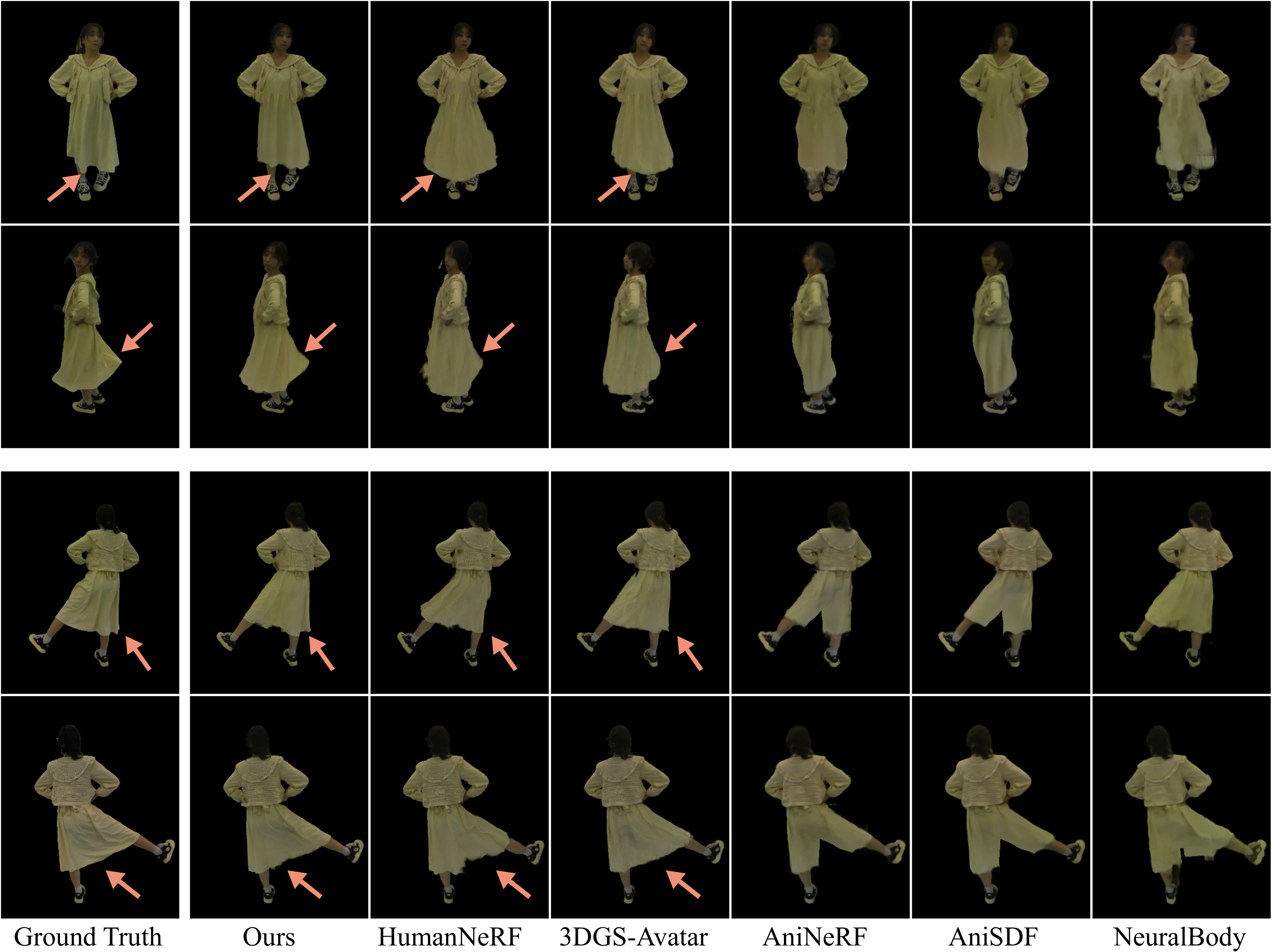}
    \caption{\textbf{Qualitative comparison on the novel pose of I3D-Human dataset.}}
    \vspace{-5mm}
    \label{fig:i3d_qualitative_cmp_novelpose}
\end{figure}

\smallskip \noindent \textbf{I3D-Human:}
We show the comparisons on I3D-Human in \cref{table:sota_I3DH,fig:i3d_qualitative_cmp_novelview,fig:i3d_qualitative_cmp_novelpose}. Our approach not only surpasses state-of-the-arts in terms of standard frame-wise similarity metrics but also achieves significantly lower DME, indicating its effectiveness in modeling inertia-induced dynamics. This improvement remains consistent across both novel view and novel pose test sets, indicating that the learned correlation between dynamic context and appearance variation in training frames can be extended to unseen poses. We will delve deeper into the generalization performance later. 


\begin{figure}[ht]
    \centering
    \includegraphics[width=0.99\textwidth]{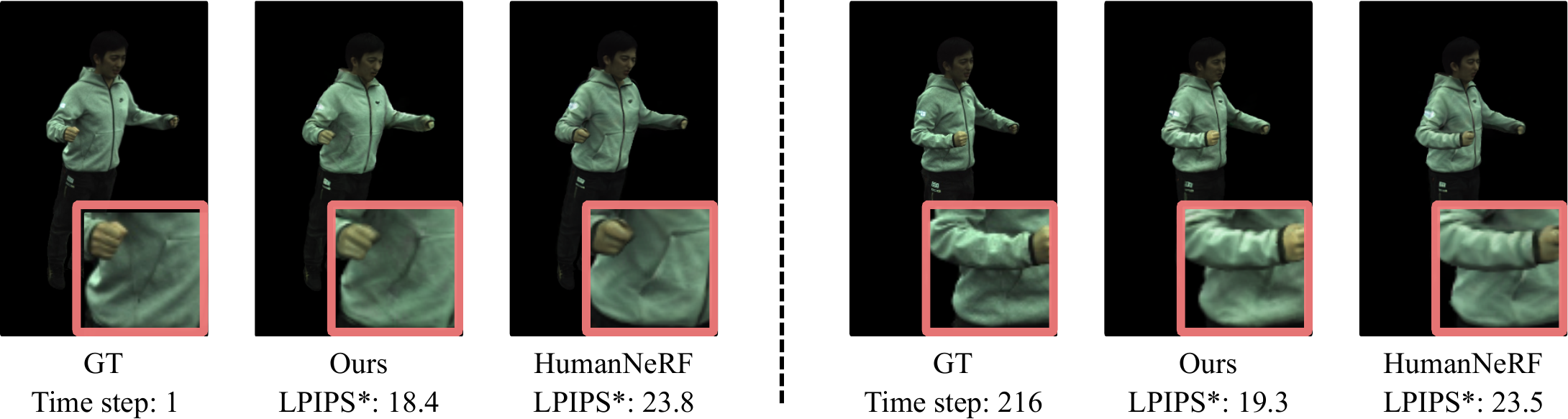} 
    \caption{The ZJU-MoCap dataset also exhibits ambiguous mapping from the static pose to various appearances. At two frames when the subject has similar poses, the clothing wrinkles can differ due to distinct past motion. Our method can reflect the variation, while HumanNeRF~\cite{weng2022humannerf} generates similar patterns.}
    \label{fig:zju_one2many}
\end{figure}

\smallskip \noindent \textbf{ZJU-MoCap:} To show Dyco maintains the capability of modeling motion with minimal inertia influence, we compare it with state-of-the-arts on ZJU-MoCap and report the metrics in~\cref{table:sota_zju}. Despite the deliberate minimization of inertia effects in human motion within ZJU-MoCap, our method achieves the best LPIPS scores in the novel-view test set. Even for ZJU-MoCap, in some training frames with similar static poses, variations in cloth wrinkles can still occur due to divergent past motion trajectories. We show that Dyco outperforms other methods in capturing the variation in~\cref{fig:zju_one2many}. In the novel pose test, our method only slightly underperforms HumanNeRF and 3DGS-Avatar. Qualitative comparison can be found in the supplementary material.

\subsection{Ablations}
\setlength\tabcolsep{.5em}
\begin{table}[t]
\centering
\caption{Ablation on pose condition $\bm{p}_i$, delta pose condition $S_{\Delta{\bm{p}}}$ and localized spatial dependency. }
\resizebox{\textwidth}{!}{
\begin{tabular}{ccccccccccc}
\toprule
 \multirow{2}{*}{\makecell{Pose\\condition}} & \multirow{2}{*}{\makecell{Delta\\condition}} &
  \multirow{2}{*}{\makecell{Local}} &  \multicolumn{4}{c}{Novel View} & 
  \multicolumn{4}{c}{Novel Pose} \\ 
  \cmidrule(lr){4-7} \cmidrule(lr){8-11} 
   & & & PSNR$\uparrow$ & SSIM$\uparrow$ & $\textrm{LPIPS}^*\downarrow$ & DME$\downarrow$ & PSNR$\uparrow$ & SSIM$\uparrow$ & $\textrm{LPIPS}^*\downarrow$ & DME$\downarrow$ \\
  \midrule  
  - & - & \checkmark 
  & 29.53 &	0.9676 & 	41.60 & 5.21	& 28.75 &	0.9639 &	45.28 & 4.45 \\
  \checkmark & - & \checkmark 
  & 30.40 &	0.9708 & 	37.97 & 4.85	& 29.16 &	0.9656 &	43.67 & 4.32 \\
  - & \checkmark & \checkmark
  & 30.89 &	0.9722 & 	36.35 & 4.56	& 29.92 &	0.9678 &	41.85 & 4.02 \\
  \checkmark & \checkmark & -
  & 30.37 &	0.9703 & 	38.78 & \textbf{4.42}	& 29.58 &	0.9666 &	42.73 & \textbf{3.77} \\
  \checkmark & \checkmark & \checkmark 
  &\textbf{ 31.22} &	\textbf{0.9738} & 	\textbf{34.54} & 4.52	& \textbf{30.12} &	\textbf{0.9691} &	\textbf{39.55} & 3.98 \\
\bottomrule
\end{tabular}}
\label{table:Results_Ablation-local-dependency}
\end{table}

\setlength\tabcolsep{.5em}
\begin{table}[h!]
\centering
\caption{Ablation on delta pose condition $S_{\Delta{\bm{p}}}$ in the non-rigid transformation and canonical volume.}

\resizebox{\textwidth}{!}{
\begin{tabular}{cccccccccc}
\toprule
 \multirow{2}{*}{$S_{\Delta{\bm{p}}}$ in $T_{\textrm{NR}}$} & \multirow{2}{*}{$S_{\Delta{\bm{p}}}$ in $F_c$} &       \multicolumn{4}{c}{Novel View} & 
  \multicolumn{4}{c}{Novel Pose} \\ 
  \cmidrule(lr){3-6} \cmidrule(lr){7-10} 
   & & PSNR$\uparrow$ & SSIM$\uparrow$ & $\textrm{LPIPS}^*\downarrow$ & DME$\downarrow$ & PSNR$\uparrow$ & SSIM$\uparrow$ & $\textrm{LPIPS}^*\downarrow$ & DME$\downarrow$ \\
  \midrule  
  - & -  
  & 29.53 &	0.9676 & 	41.60 & 5.21	& 28.75 &	0.9639 &	45.28 & 4.45\\
  \checkmark & -  
  & 31.13 &	0.9729 & 	35.86 & 5.14	& 29.99 &	0.9678 &	41.52 & 4.26 \\
  \checkmark & \checkmark
  & \textbf{31.22} &	\textbf{0.9738} & 	\textbf{34.54} & \textbf{4.52}	& \textbf{30.12} &	\textbf{0.9691} &	\textbf{39.55} & \textbf{3.98} \\

\bottomrule
\end{tabular}}
\label{table: dpose in ca}

\end{table}

\noindent \textbf{Pose condition $\bm{p}_i$} provides global spatial information for the current frame during training, and its removal results in decreased performance, see ~\cref{table:Results_Ablation-local-dependency}.

\noindent \textbf{Delta pose condition $S_{\Delta{\bm{p}}}$} incorporates temporal human inertia knowledge, which we believe is the core design for resolving ambiguities related to dynamic context. Discarding this component prevents the network from leveraging inertia information and degrades the performance, see~\cref{table:Results_Ablation-local-dependency}.

\noindent \textbf{Localization} in dynamic context encoder reduces the complexity of $S_{\Delta{\bm{p}}}$ input by masking redundant dependencies. Without this module, the overfitted network will struggle in generalizing to novel poses, see~\cref{table:Results_Ablation-local-dependency}.

\noindent \textbf{Conditional input to $T_\textrm{NR}$ and $F_c$:} In \cref{sec:Ambiguities}, we analyze two types of dynamic context related ambiguities and propose to input delta pose $S_{\Delta{\bm{p}}}$ to both non-rigid transformation and canonical volume. \cref{table: dpose in ca} validates that inputting $S_{\Delta{\bm{p}}}$ into both modules helps resolve the ambiguities.

\begin{figure}[t]
    \centering
	\includegraphics[width=.6\columnwidth]{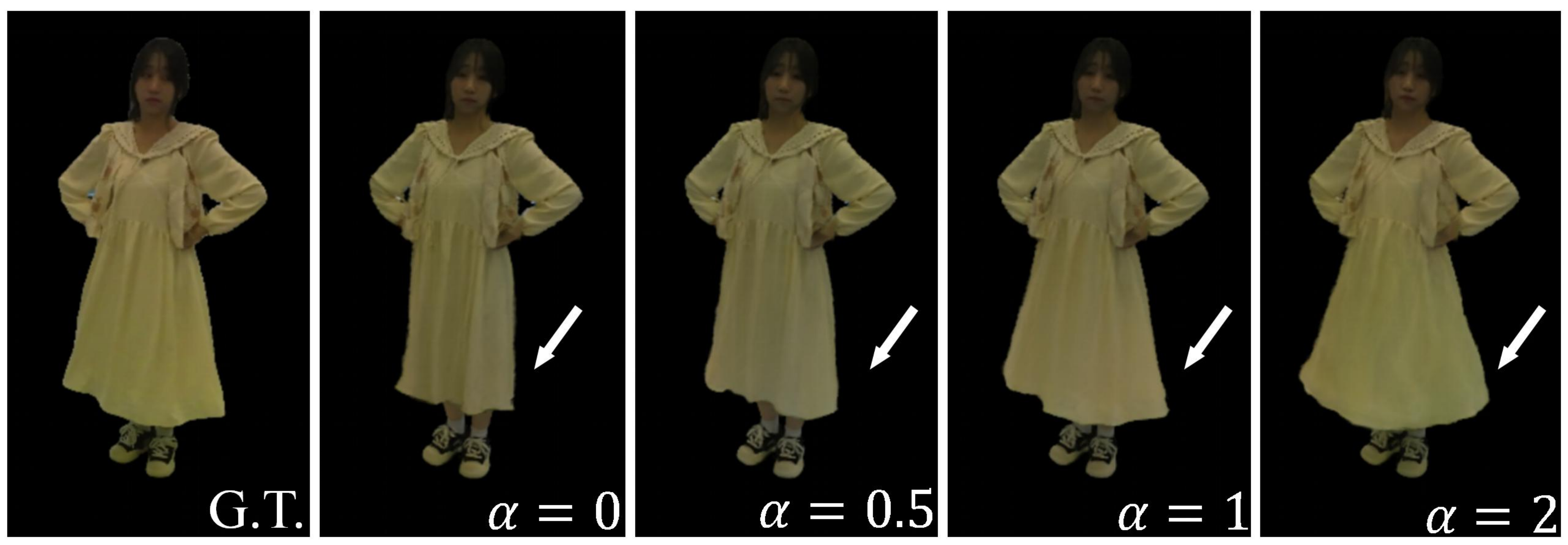}
	\caption{Appearance variation under varying spinning velocity. We input the scaled delta-sequence $\alpha S_{\Delta{\bm{p}}}$ to the trained model to obtain different renderings.}
	\label{fig:novel-velocity}
\end{figure}

\begin{figure}[t]
    \centering
	\includegraphics[width=.9\columnwidth]{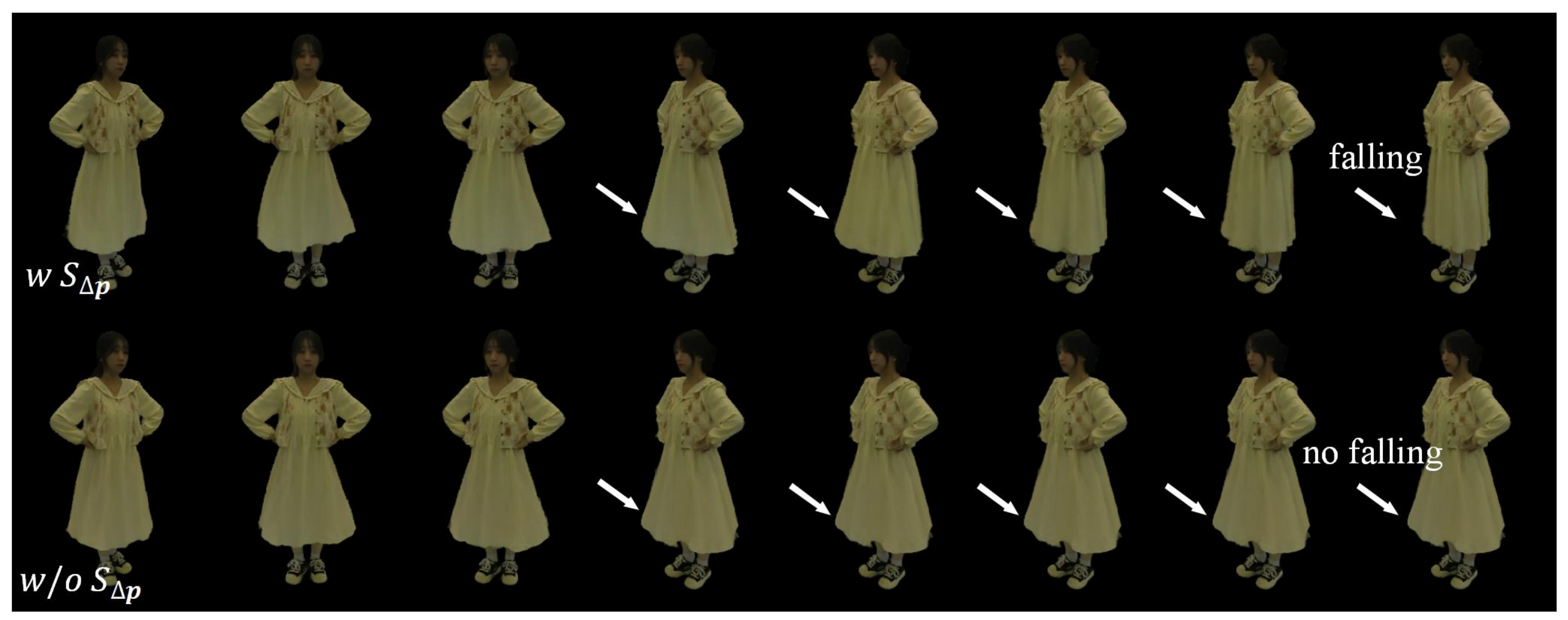}
	\caption{Dyco correctly models the falling of the drape when the character stops abruptly during spinning, whereas the baseline keeps the drape still.}
	\label{fig:novel-acceleration}
\end{figure}

\subsection{Generalization to Novel Dynamic Context}
\label{subsec:Analysis on Novel Pose Generalization}
To showcase that Dyco correctly models the correlation between inertia factors and appearance variations, we create novel dynamic context by varying the velocity of the training poses to study how it impacts the rendered results. Please refer to the supplemental video for a clearer visualization.

\noindent \textbf{Varying velocity:} We scale $S_{\Delta{\bm{p}}}$ with a factor $\alpha$ ranging from $0$ to $2$, where $\alpha>1$ speeds up motion and use $\alpha{S_{\Delta{\bm{p}}}}$ as the sequence condition. \cref{fig:novel-velocity} demonstrates Dyco's ability to accurately portray varying drape lifting amplitudes with different spinning speeds, with larger $\alpha$ yielding greater amplitudes.

\noindent \textbf{Varying acceleration:} We simulate an abrupt stop effect by interrupting poses midway and making subsequent poses identical. The modified pose sequence is then used as the input condition. As shown in \cref{fig:novel-acceleration}, Dyco, when conditioned on the novel pose sequence, accurately animates the falling of the hem. In contrast, the model without pose sequence inputs fails to depict this motion.


\section{Conclusion}

In this work, we introduce Dyco, a novel human motion modeling method that incorporates pose-sequence conditions to address appearance ambiguities resulting from different dynamic contexts. We argue that human appearance is influenced not only by pose conditions but also by past motion trajectories, which can be adequately captured by pose sequences. To mitigate overfitting caused by excessive reliance on delta pose, we design a localized dynamic context encoder. This approach allows us to resolve appearance ambiguities caused by dynamic context and improve the rendering quality of human bodies in loose attire. The I3D-Human dataset addresses previous oversights in datasets and advances research into real-life human motion.
\section*{Acknowledgements}
This work is supported by the Shanghai Artificial intelligence Laboratory, in part by JSPS KAKENHI Grant Numbers 24K22318, 22H00529, 20H05951, JST-Mirai Program JPMJMI23G1.

\par\vfill\par
\clearpage  

%
%
\bibliographystyle{splncs04}
\bibliography{main}

\begin{thebibliography}{10}
\providecommand{\url}[1]{\texttt{#1}}
\providecommand{\urlprefix}{URL }
\providecommand{\doi}[1]{https://doi.org/#1}

\bibitem{easymocap}
Easymocap - make human motion capture easier. Github (2021), \url{https://github.com/zju3dv/EasyMocap}

\bibitem{loper2023smpl}
{SMPL: A skinned multi-person linear model}, author={Loper, Matthew and Mahmood, Naureen and Romero, Javier and Pons-Moll, Gerard and Black, Michael J}. In: Seminal Graphics Papers: Pushing the Boundaries, Volume 2, pp. 851--866 (2023)

\bibitem{Alldieck2018_PeopleSnapShot}
Alldieck, T., Magnor, M.A., Xu, W., Theobalt, C., Pons{-}Moll, G.: Video based reconstruction of 3d people models. In: 2018 {IEEE} Conference on Computer Vision and Pattern Recognition, {CVPR} 2018, Salt Lake City, UT, USA, June 18-22, 2018. pp. 8387--8397. Computer Vision Foundation / {IEEE} Computer Society (2018). \doi{10.1109/CVPR.2018.00875}, \url{http://openaccess.thecvf.com/content\_cvpr\_2018/html/Alldieck\_Video\_Based\_Reconstruction\_CVPR\_2018\_paper.html}

\bibitem{anguelov2005scape}
Anguelov, D., Srinivasan, P., Koller, D., Thrun, S., Rodgers, J., Davis, J.: {Scape: shape completion and animation of people}. In: ACM SIGGRAPH 2005 Papers, pp. 408--416 (2005)

\bibitem{Chan2022_Eg3D}
Chan, E.R., Lin, C.Z., Chan, M.A., Nagano, K., Pan, B., Mello, S.D., Gallo, O., Guibas, L.J., Tremblay, J., Khamis, S., Karras, T., Wetzstein, G.: Efficient geometry-aware 3d generative adversarial networks. In: {IEEE/CVF} Conference on Computer Vision and Pattern Recognition, {CVPR} 2022, New Orleans, LA, USA, June 18-24, 2022. pp. 16102--16112. {IEEE} (2022). \doi{10.1109/CVPR52688.2022.01565}, \url{https://doi.org/10.1109/CVPR52688.2022.01565}

\bibitem{chen2022tensorf}
Chen, A., Xu, Z., Geiger, A., Yu, J., Su, H.: {Tensorf: Tensorial radiance fields}. In: European Conference on Computer Vision. pp. 333--350. Springer (2022)

\bibitem{cheng2023dna}
Cheng, W., Chen, R., Fan, S., Yin, W., Chen, K., Cai, Z., Wang, J., Gao, Y., Yu, Z., Lin, Z., et~al.: {Dna-rendering: A diverse neural actor repository for high-fidelity human-centric rendering}. In: Proceedings of the IEEE/CVF International Conference on Computer Vision. pp. 19982--19993 (2023)

\bibitem{dong2021fast}
Dong, J., Fang, Q., Jiang, W., Yang, Y., Bao, H., Zhou, X.: Fast and robust multi-person 3d pose estimation and tracking from multiple views. In: T-PAMI (2021)

\bibitem{Freifeld_LieBodies}
Freifeld, O., Black, M.J.: Lie bodies: {A} manifold representation of 3d human shape. In: Fitzgibbon, A.W., Lazebnik, S., Perona, P., Sato, Y., Schmid, C. (eds.) Computer Vision - {ECCV} 2012 - 12th European Conference on Computer Vision, Florence, Italy, October 7-13, 2012, Proceedings, Part {I}. Lecture Notes in Computer Science, vol.~7572, pp. 1--14. Springer (2012). \doi{10.1007/978-3-642-33718-5\_1}, \url{https://doi.org/10.1007/978-3-642-33718-5\_1}

\bibitem{guan2012drape}
Guan, P., Reiss, L., Hirshberg, D.A., Weiss, A., Black, M.J.: Drape: Dressing any person. ACM Transactions on Graphics (ToG)  \textbf{31}(4),  1--10 (2012)

\bibitem{Hirshberg_SCAPE}
Hirshberg, D.A., Loper, M., Rachlin, E., Black, M.J.: Coregistration: Simultaneous alignment and modeling of articulated 3d shape. In: Fitzgibbon, A.W., Lazebnik, S., Perona, P., Sato, Y., Schmid, C. (eds.) Computer Vision - {ECCV} 2012 - 12th European Conference on Computer Vision, Florence, Italy, October 7-13, 2012, Proceedings, Part {VI}. Lecture Notes in Computer Science, vol.~7577, pp. 242--255. Springer (2012). \doi{10.1007/978-3-642-33783-3\_18}, \url{https://doi.org/10.1007/978-3-642-33783-3\_18}

\bibitem{Ionescu2014_Human3.6M}
Ionescu, C., Papava, D., Olaru, V., Sminchisescu, C.: Human3.6m: Large scale datasets and predictive methods for 3d human sensing in natural environments. {IEEE} Trans. Pattern Anal. Mach. Intell.  \textbf{36}(7),  1325--1339 (2014). \doi{10.1109/TPAMI.2013.248}, \url{https://doi.org/10.1109/TPAMI.2013.248}

\bibitem{jiang2023instantavatar}
Jiang, T., Chen, X., Song, J., Hilliges, O.: Instantavatar: Learning avatars from monocular video in 60 seconds. In: Proceedings of the IEEE/CVF Conference on Computer Vision and Pattern Recognition. pp. 16922--16932 (2023)

\bibitem{kajiya1984ray}
Kajiya, J.T., Von~Herzen, B.P.: {Ray Tracing Volume Densities}. ACM SIGGRAPH computer graphics  \textbf{18}(3),  165--174 (1984)

\bibitem{kerbl20233d}
Kerbl, B., Kopanas, G., Leimk{\"u}hler, T., Drettakis, G.: 3d gaussian splatting for real-time radiance field rendering. ACM Transactions on Graphics  \textbf{42}(4) (2023)

\bibitem{kerbl2023_3dgs}
Kerbl, B., Kopanas, G., Leimk{\"{u}}hler, T., Drettakis, G.: 3d gaussian splatting for real-time radiance field rendering. {ACM} Trans. Graph.  \textbf{42}(4),  139:1--139:14 (2023). \doi{10.1145/3592433}, \url{https://doi.org/10.1145/3592433}

\bibitem{Kingma_Adam}
Kingma, D.P., Ba, J.: Adam: {A} method for stochastic optimization. In: Bengio, Y., LeCun, Y. (eds.) 3rd International Conference on Learning Representations, {ICLR} 2015, San Diego, CA, USA, May 7-9, 2015, Conference Track Proceedings (2015), \url{http://arxiv.org/abs/1412.6980}

\bibitem{Larboulette_DynamicSkinning}
Larboulette, C., Cani, M.P., Arnaldi, B.: Dynamic skinning: adding real-time dynamic effects to an existing character animation. In: Proceedings of the 21st Spring Conference on Computer Graphics. p. 87–93. SCCG '05, Association for Computing Machinery, New York, NY, USA (2005). \doi{10.1145/1090122.1090138}, \url{https://doi.org/10.1145/1090122.1090138}

\bibitem{lewis2023pose}
Lewis, J.P., Cordner, M., Fong, N.: Pose space deformation: a unified approach to shape interpolation and skeleton-driven deformation. In: Seminal Graphics Papers: Pushing the Boundaries, Volume 2, pp. 811--818 (2023)

\bibitem{li2022tava}
Li, R., Tanke, J., Vo, M., Zollhofer, M., Gall, J., Kanazawa, A., Lassner, C.: Tava: Template-free animatable volumetric actors (2022)

\bibitem{lin2022efficient}
Lin, H., Peng, S., Xu, Z., Yan, Y., Shuai, Q., Bao, H., Zhou, X.: {Efficient neural radiance fields for interactive free-viewpoint video}. In: SIGGRAPH Asia 2022 Conference Papers. pp.~1--9 (2022)

\bibitem{Liu2021_NeuralActor}
Liu, L., Habermann, M., Rudnev, V., Sarkar, K., Gu, J., Theobalt, C.: Neural actor: neural free-view synthesis of human actors with pose control. ACM Trans. Graph.  \textbf{40}(6) (dec 2021). \doi{10.1145/3478513.3480528}, \url{https://doi.org/10.1145/3478513.3480528}

\bibitem{mildenhall2021nerf}
Mildenhall, B., Srinivasan, P.P., Tancik, M., Barron, J.T., Ramamoorthi, R., Ng, R.: {NeRF: Representing Scenes as Neural Radiance Fields for View Synthesis}. Communications of the ACM  \textbf{65}(1),  99--106 (2021)

\bibitem{muller2022instant}
M{\"u}ller, T., Evans, A., Schied, C., Keller, A.: Instant neural graphics primitives with a multiresolution hash encoding. ACM Transactions on Graphics (ToG)  \textbf{41}(4),  1--15 (2022)

\bibitem{noguchi2021neural}
Noguchi, A., Sun, X., Lin, S., Harada, T.: Neural articulated radiance field. In: Proceedings of the IEEE/CVF International Conference on Computer Vision. pp. 5762--5772 (2021)

\bibitem{patel20_tailornet}
Patel, C., Liao, Z., Pons-Moll, G.: Tailornet: Predicting clothing in 3d as a function of human pose, shape and garment style. In: {IEEE} Conference on Computer Vision and Pattern Recognition (CVPR). {IEEE} (jun 2020)

\bibitem{peng2021animatable}
Peng, S., Dong, J., Wang, Q., Zhang, S., Shuai, Q., Zhou, X., Bao, H.: Animatable neural radiance fields for modeling dynamic human bodies. In: ICCV (2021)

\bibitem{peng2023implicit}
Peng, S., Geng, C., Zhang, Y., Xu, Y., Wang, Q., Shuai, Q., Zhou, X., Bao, H.: Implicit neural representations with structured latent codes for human body modeling. IEEE Transactions on Pattern Analysis and Machine Intelligence  (2023)

\bibitem{peng2024animatable}
Peng, S., Xu, Z., Dong, J., Wang, Q., Zhang, S., Shuai, Q., Bao, H., Zhou, X.: Animatable implicit neural representations for creating realistic avatars from videos. TPAMI  (2024)

\bibitem{peng2021_neuralbody}
Peng, S., Zhang, Y., Xu, Y., Wang, Q., Shuai, Q., Bao, H., Zhou, X.: Neural body: Implicit neural representations with structured latent codes for novel view synthesis of dynamic humans. In: {IEEE} Conference on Computer Vision and Pattern Recognition, {CVPR} 2021, virtual, June 19-25, 2021. pp. 9054--9063. Computer Vision Foundation / {IEEE} (2021). \doi{10.1109/CVPR46437.2021.00894}, \url{https://openaccess.thecvf.com/content/CVPR2021/html/Peng\_Neural\_Body\_Implicit\_Neural\_Representations\_With\_Structured\_Latent\_Codes\_for\_CVPR\_2021\_paper.html}

\bibitem{peng2021neural}
Peng, S., Zhang, Y., Xu, Y., Wang, Q., Shuai, Q., Bao, H., Zhou, X.: Neural body: Implicit neural representations with structured latent codes for novel view synthesis of dynamic humans. In: CVPR (2021)

\bibitem{PonsMoll2015_Dyna}
Pons{-}Moll, G., Romero, J., Mahmood, N., Black, M.J.: Dyna: a model of dynamic human shape in motion. {ACM} Trans. Graph.  \textbf{34}(4),  120:1--120:14 (2015). \doi{10.1145/2766993}, \url{https://doi.org/10.1145/2766993}

\bibitem{qian2023_3dgsavatar}
Qian, Z., Wang, S., Mihajlovic, M., Geiger, A., Tang, S.: 3dgs-avatar: Animatable avatars via deformable 3d gaussian splatting  (2024)

\bibitem{Rohmer2021VelocitySF}
Rohmer, D., Tarini, M., Kalyanasundaram, N., Moshfeghifar, F., Cani, M.P., Zordan, V.B.: Velocity skinning for real‐time stylized skeletal animation. Computer Graphics Forum  \textbf{40} (2021), \url{https://api.semanticscholar.org/CorpusID:233210320}

\bibitem{shuai2022multinb}
Shuai, Q., Geng, C., Fang, Q., Peng, S., Shen, W., Zhou, X., Bao, H.: Novel view synthesis of human interactions from sparse multi-view videos. In: SIGGRAPH Conference Proceedings (2022)

\bibitem{Yu2021_ANeRF}
Su, S., Yu, F., Zollh{\"{o}}fer, M., Rhodin, H.: A-nerf: Articulated neural radiance fields for learning human shape, appearance, and pose. In: Ranzato, M., Beygelzimer, A., Dauphin, Y.N., Liang, P., Vaughan, J.W. (eds.) Advances in Neural Information Processing Systems 34: Annual Conference on Neural Information Processing Systems 2021, NeurIPS 2021, December 6-14, 2021, virtual. pp. 12278--12291 (2021), \url{https://proceedings.neurips.cc/paper/2021/hash/65fc9fb4897a89789352e211ca2d398f-Abstract.html}

\bibitem{Teed2020_RAFT}
Teed, Z., Deng, J.: {RAFT:} recurrent all-pairs field transforms for optical flow. In: Vedaldi, A., Bischof, H., Brox, T., Frahm, J. (eds.) Computer Vision - {ECCV} 2020 - 16th European Conference, Glasgow, UK, August 23-28, 2020, Proceedings, Part {II}. Lecture Notes in Computer Science, vol. 12347, pp. 402--419. Springer (2020). \doi{10.1007/978-3-030-58536-5\_24}, \url{https://doi.org/10.1007/978-3-030-58536-5\_24}

\bibitem{wang2021neus}
Wang, P., Liu, L., Liu, Y., Theobalt, C., Komura, T., Wang, W.: Neus: Learning neural implicit surfaces by volume rendering for multi-view reconstruction. arXiv preprint arXiv:2106.10689  (2021)

\bibitem{wang2022arah}
Wang, S., Schwarz, K., Geiger, A., Tang, S.: Arah: Animatable volume rendering of articulated human sdfs. In: European conference on computer vision. pp. 1--19. Springer (2022)

\bibitem{weng2022humannerf}
Weng, C.Y., Curless, B., Srinivasan, P.P., Barron, J.T., Kemelmacher-Shlizerman, I.: Human{N}e{RF}: Free-viewpoint rendering of moving people from monocular video. In: Proceedings of the IEEE/CVF Conference on Computer Vision and Pattern Recognition (CVPR). pp. 16210--16220 (June 2022)

\bibitem{wu2020multi}
Wu, M., Wang, Y., Hu, Q., Yu, J.: {Multi-view neural human rendering}. In: Proceedings of the IEEE/CVF Conference on Computer Vision and Pattern Recognition. pp. 1682--1691 (2020)

\bibitem{yu2023monohuman}
Yu, Z., Cheng, W., Liu, X., Wu, W., Lin, K.Y.: Monohuman: Animatable human neural field from monocular video. In: Proceedings of the IEEE/CVF Conference on Computer Vision and Pattern Recognition. pp. 16943--16953 (2023)

\bibitem{zhang2018_lpips}
Zhang, R., Isola, P., Efros, A.A., Shechtman, E., Wang, O.: The unreasonable effectiveness of deep features as a perceptual metric. In: CVPR (2018)

\bibitem{zheng_AvatarRex2023}
Zheng, Z., Zhao, X., Zhang, H., Liu, B., Liu, Y.: Avatarrex: Real-time expressive full-body avatars. ACM Trans. Graph.  \textbf{42}(4) (jul 2023). \doi{10.1145/3592101}, \url{https://doi.org/10.1145/3592101}

\end{thebibliography}

\newpage

\setcounter{section}{0}
\section*{\Large Supplementary Material}
Due to the limited space of the main text, we provide a detailed description of our method implementation, the proposed I3D-Human dataset, results of more ablation studies and on ZJU-MoCap, and also discuss our limitations.

\section{Implementation Details}

\subsection{Localized Dynamic Context Encoding:} 

 Our localized dynamic context encoding $\Phi_{\textrm{seq}}(S_\Delta{{\bm{p}}})$ consists of three stages, as shown in Fig. 2 of the main paper. We instantiate two localized dynamic context encoders of the same architecture to produce conditions for the rigid transformation and non-rigid transformation modules. The output condition vectors are concatenated with the sinusoidal positional encodings of the input coordinate to serve as the inputs to the subsequent transformation modules.

\smallskip \noindent \textbf{Kinematically-guided Spatial Dependency:} For a given context sequence $S_{\Delta{\bm{p}}}$ and an input coordinate $\bm{x}$, we begin by processing each $\Delta{\bm{p}}\in \mathbb{R}^{3\times K+3}$ independently within the sequence. Here we leverage a reasonable physical assumption: the impact of a rotating joint is mainly restricted to the kinematic chains to which it belongs. For instance, the movement of the hips generally has minimal impact on the appearance of the shoulders. Therefore, the motion and appearance of a point $\bm{x}$ is expected to mainly depend on the joints within the same kinematic chains.  With an input coordinate $\bm{x}$ and its associated blending weights $W_c(\bm{x})\in \mathbb{R}^{K}$, we first identify its nearest joint $k=\argmax W_c(\bm{x})$ to determine the kinematic chains $\bm{x}$ lies on. We use the kinematic chains predefined in the skeleton model used by SMPL. Then we gather all joints from the related kinematic chains. Last, we apply a binary mask to each $\Delta{\bm{p}}$ to filter out the unrelated joints that fall outside the kinematic chains. We denote masked conditions as $\hat{\Delta}{\bm{p}}$. We describe the effectiveness of this design in~\cref{sec:local-dependency}.

\smallskip \noindent \textbf{Spatial Encoding:} We then flatten the masked spatial condition $\hat{\Delta}{\bm{p}}$ to an 1D vector and project it to a $16$-dim vector by an one-layered MLP with ReLU activation. We pass the vector of each time step through the same MLP, resulting in a sequence of $16$-dimensional vectors with a length of $L$.

\smallskip \noindent \textbf{Temporal Encoding:} We concatenate the spatially-encoded sequence into an 1D vector, then pass it through a second one-layered MLP with ReLU activation to reduce it to a 32-dimensional vector. This temporally-aggregated vector serves as the localized dynamic context condition for the query coordinate $\bm{x}$ and is denoted as $\Phi_{seq}(S_{\Delta{\bm{p}}})$. For simplicity, we omit $\bm{x}$ here as it only affects the masking stage.

\subsection{Other Modules:}
\smallskip \noindent \textbf{Canonical Volume:} We implement the canonical volume $F_c$ as a tri-plane encoder~\cite{chen2022tensorf}, followed by an MLP with a color branch and a density branch. As for the tri-plane grids, we use multi-scale planes with $4$ different resolutions at $64^2$, $128^2$, $256^2$, and $512^2$. The per-plane and per-scale features are multiplied across planes and concatenated across scales. We set each per-plane and per-scale dimension as $32$ so the final encoded feature has dimension $128$. The Color MLP has 2 hidden layers with a channel dimension as $256$ and the density branch has a single hidden layer with a channel dimension of $256$.

\smallskip \noindent \textbf{Rigid Transformation Module:} We adopt the implementation in HumanNeRF~\cite{weng2022humannerf}. In specific, the world coordinate $\bm{x}$ is rigidly transformed to an approximate canonical coordinate $\bm{x}_r$ by a backward LBS as Eq.(4) of the main text. The backward linear blend weights are estimated by the softmax-normalized grid weights of transformed $\bm{x}$ as Eq.(5) of the main text. The volume grid is a 3D CNN with a volume size of $32^3$. Please refer to HumanNeRF~\cite{weng2022humannerf} for more details.

\smallskip \noindent \textbf{Non-Rigid Transformation Module:} Our non-rigid transformation module is a two-layered MLP with width=128. The input coordinate is embedded as sinusoidal positional encodings~\cite{mildenhall2021nerf} with a frequency level of 6, which is then concatenated with the dynamic context condition before feeding to the MLP.

\smallskip \noindent \textbf{Pose-sequence Related Input:} For the delta-pose sequence condition 
$S_{\Delta{\bm{p}}}=\{\Delta{\bm{p}_{i-(L-1)s}},\ldots,\Delta{\bm{p}_i}\}$. 
We set $L=6$, $s = 25$. So the context covers the motion history of 150 frames forward.  We embed the angles rotating the joints from the last step to the current step and represent them in the axis-angle form. As the pose difference between two consecutive frames can be small and noisy, we take a stride $s_d=25$ backwards and compute the delta as $\Delta{\bm{p}_i} = \delta(\bm{p}_i, \bm{p}_{i-s_d})$. Additionally, we include a 3-dimensional global translation vector and concatenate it with the joint rotation vectors, yielding an $L$-length sequence of delta vectors with a dimension of $3K+3$, where $K$ is the number of joints.

\smallskip \noindent \textbf{Optimization:} We jointly train the whole network using both $\mathcal{L}_{\textrm{LPIPS}}$ and $\mathcal{L}_{\textrm{MSE}}$ and set the weights as $0.2:1$. 
We use the Adam optimizer~\cite{Kingma_Adam} with $\beta_1=0.9, \beta_2=0.99$ and set the learning rate as 5e-4 for the canonical triplane module and 5e-5 for all the other modules. We train the models on a single GeForce RTX 2080 Ti GPU for 200k iterations which takes about 12 hours.

\section{Dynamic Motion Error}
\begin{figure}[t!]
    \centering
    \includegraphics[width=0.9\textwidth]{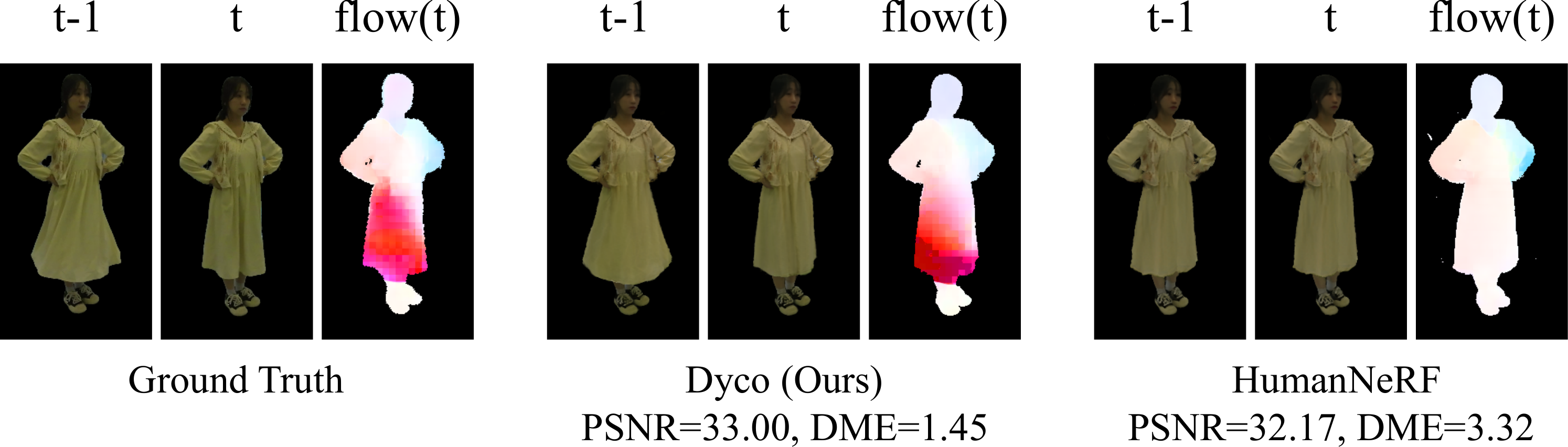}
    \caption{DME, \emph{The lower the better}, focuses on evaluating the rendering of dynamic motion effects such as the gentle falling of the drape. The flow(t) is the estimated optical flow that warps pixels in rendering t back to rendering t-1. We show the PSNR and DME of the rendering at step t.}
    \label{fig:dme_cmp}
\end{figure}
To accurately evaluate the portrayal of dynamic motion, we introduce a new metric named Dynamic Motion Error (DME) in Sec. 4.4 of the main paper. To demonstrate the validity of this metric, we provide a visualization in~\cref{fig:dme_cmp}. We illustrate how the estimated optical flow captures the soft cascading of the drape in the ground truth images for two frames of an identical static pose. Our method accurately renders the falling effect, resulting in a small error in optical flow estimation (DME=1.45) compared to the ground truth. In contrast, HumanNeRF produces similar renderings at the two steps, depicting the dress drape as motionless or static. Consequently, there is a significant discrepancy between its flow estimation and that of the ground truth (DME=3.32). It's worth noting that, compared to per-frame metrics such as PSNR, the motion-based DME provides a better reflection of the performance gap between models in its absolute value.

\section{I3D-Human Dataset}
\begin{figure}[h!]
    \begin{minipage}[t]{1.\textwidth}
        \centering
        \captionof{table}{Details of our I3D-Human Dataset, including available views, training views, novel view split and novel pose split.}
        \resizebox{1.\textwidth}{!}{
        \begin{tabular}{ccccccccc}
        \toprule
         {Sequence} &  Description & 
          Available Views & Training Views & Novel View Split & Novel Pose Split\\ 
        \midrule
        ID1\_1 &  Spinning in dress & [1, 2, 3, 4, 6, 7, 8, 9] & [1, 3, 6, 8] & [301, 2301] & [2301, 2600] \\  
        ID1\_2 &  Swinging legs in dress & [0, 1, 2, 5, 6, 7, 8, 9] & [0, 2, 5, 7] & [301, 2301] & [2301, 3800] \\  
        ID2\_1 &  Spinning in coat & [0, 1, 2, 3, 4, 5, 6, 7, 9] & [1, 3, 5, 7, 9] & [1404, 2484] & [2484, 3111] \\ 
        ID2\_2 &  Jumping in coat & [0, 1, 2, 3, 4, 5, 6, 7, 8, 9] & [1, 3, 5, 7, 9] & [1002, 2196] & [2196, 3798] \\ 
        ID3\_1 &  Spinning in coat & [0, 1, 2, 3, 4, 5, 6, 7, 8, 9] & [1, 3, 5, 7, 9] & [1407, 3255] & [3255, 3639] \\ 
        ID3\_2 &  Jumping in coat & [0, 1, 2, 3, 4, 5, 6, 7, 8, 9] & [1, 3, 5, 7, 9] & [909, 1980] & [2136, 3198] \\ 
        \bottomrule
        \end{tabular}}
        \label{tab: details of I3D-Human}
    \end{minipage}
    \begin{minipage}[t]{1.\textwidth}
        \centering
        \includegraphics[width=0.9\textwidth]{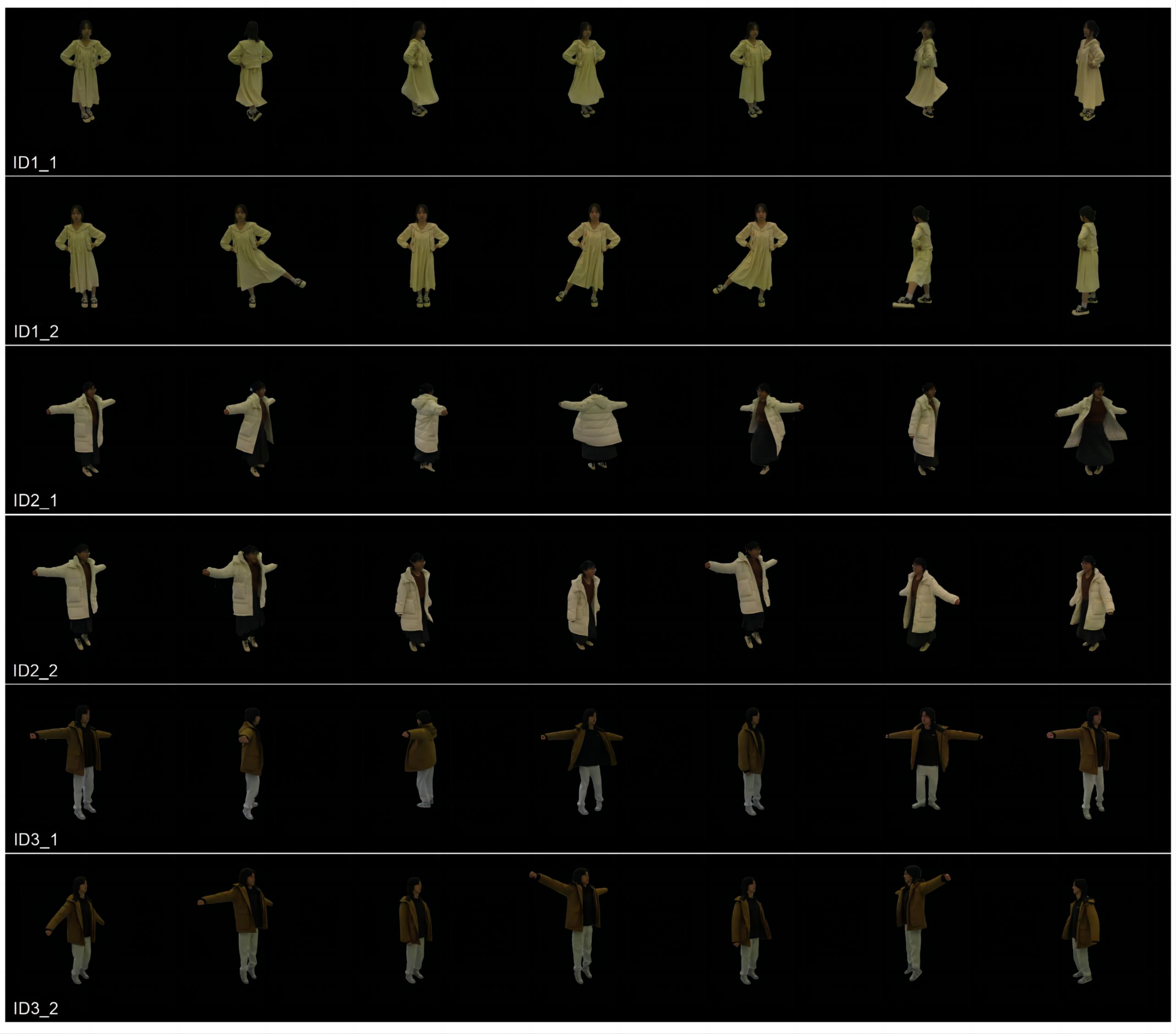}
        \captionof{figure}{Visualization of our I3D-Human dataset. Our dataset focuses on capturing variations in clothing appearance under approximately identical poses. The dataset is released for research purposes.}
        \label{fig:dataset}
    \end{minipage}
\end{figure}

I3D-Human comprises 6  multi-view sequences from different individuals, reflecting complex inertial characteristics. We show the data statistics in~\cref{tab: details of I3D-Human} and some examples in~\cref{fig:dataset}.

\section{Results on ZJU-MoCap}
Besides the quantitative comparison in Tab. 2 of the main paper, we show the qualitative comparison on the ZJU-MoCap dataset in~\cref{fig:zju_cmp}. In novel-view test set, our method can capture more details on the clothing wrinkles. Although NeuralBody~\cite{peng2023implicit} attains the highest PSNR and SSIM scores in novel views, it suffers from significant blurriness. The novel pose renderings generated by our method are quite comparable to those produced by HumanNeRF~\cite{weng2022humannerf} and 3DGS-Avatar~\cite{qian2023_3dgsavatar}.

\begin{table}[t]
\setlength\tabcolsep{.5em}
    \begin{subtable}[t]{1.\textwidth}
    \centering
    \resizebox{0.9\textwidth}{!}{
    \begin{tabular}{ccccccccc}
    \toprule
     \multirow{2}{*}{Input} &  \multicolumn{4}{c}{Novel View} & 
      \multicolumn{4}{c}{Novel Pose} \\ 
      \cmidrule(lr){2-5} \cmidrule(lr){6-9}  
       & PSNR$\uparrow$ & SSIM$\uparrow$ & $\textrm{LPIPS}^*\downarrow$ & DME$\downarrow$ & PSNR$\uparrow$ & SSIM$\uparrow$ & $\textrm{LPIPS}^*\downarrow$ & DME$\downarrow$ \\  
    \midrule
    $S_{\bm{p}}$ & 29.93 & 0.9688 & 41.63 & 5.13 & 29.22 & 0.9649 & 45.85 & 4.33 \\  
    $S_{\Delta{\bm{p}}}$ & \textbf{31.22} & \textbf{0.9738} & \textbf{34.54} & \textbf{4.52} & \textbf{30.12} & \textbf{0.9691} & \textbf{39.55} & \textbf{3.98} \\  
    \bottomrule
    \end{tabular}}
    \caption{Inputting $S_{\Delta{\bm{p}}}$ outperforms $S_{\bm{p}}$. $\textrm{LPIPS}^*=\textrm{LPIPS}\times10^3$}
    \label{table:Ablation_delta}
    \end{subtable}


    \begin{subtable}[t]{1.\textwidth}
    \centering
    \resizebox{0.9\textwidth}{!}{
    \begin{tabular}{ccccccccc}
    \toprule
     \multirow{2}{*}{Representation} &  \multicolumn{4}{c}{Novel View} & 
      \multicolumn{4}{c}{Novel Pose} \\ 
      \cmidrule(lr){2-5} \cmidrule(lr){6-9}  
       & PSNR$\uparrow$ & SSIM$\uparrow$ & $\textrm{LPIPS}^*\downarrow$ & DME$\downarrow$ & PSNR$\uparrow$ & SSIM$\uparrow$ & $\textrm{LPIPS}^*\downarrow$ & DME$\downarrow$ \\  
    \midrule
    Rodrigues & 31.20 & 0.9737 & 35.61 & 4.65 & 30.06 & 0.9688 & 40.70 & \textbf{3.88} \\
    Quaternion & 31.17 & 0.9734 & 36.00 & 4.55 & 30.07 & 0.9685 & 41.07 & 3.97 \\
    Axis-angle & \textbf{31.22} & \textbf{0.9738} & \textbf{34.54} & \textbf{4.52} & \textbf{30.12} & \textbf{0.9691} & \textbf{39.55} & 3.98 \\  
    \bottomrule
    \end{tabular}}
    \caption{The impact of the delta pose representation.}
    \label{table:Ablation_rot-rep}
    \end{subtable}

    \begin{subtable}[t]{1.\textwidth}
    \centering
    \resizebox{0.9\textwidth}{!}{
    \begin{tabular}{cccccccccc}
    \toprule
     \multirow{2}{*}{$L_{d}$} &\multirow{2}{*}{$s \,\& \,s_d$} &  \multicolumn{4}{c}{Novel View} & 
      \multicolumn{4}{c}{Novel Pose} \\ 
      \cmidrule(lr){3-6} \cmidrule(lr){7-10}  
       && PSNR$\uparrow$ & SSIM$\uparrow$ & $\textrm{LPIPS}^*\downarrow$ & DME$\downarrow$ & PSNR$\uparrow$ & SSIM$\uparrow$ & $\textrm{LPIPS}^*\downarrow$ & DME$\downarrow$ \\  
    \midrule
    2 &25& 31.04 & 0.9733 & 35.05 & \textbf{4.50} & 29.89 & 0.9685 & 40.08 & 3.94 \\
    6 &12& 30.99 & 0.9726 & 36.34 & 4.78 & 29.95 & 0.9682 & 41.31 & 4.04 \\ 
    6(default) &25(default)& \textbf{31.22} & \textbf{0.9738} & \textbf{34.54} & 4.52 & \textbf{30.12} & \textbf{0.9691} & \textbf{39.55} & 3.98 \\  
    12 &25& 31.21 & 0.9734 & 35.14 & 4.6 & 29.69 & 0.9672 & 42.53 & \textbf{3.91} \\ 
    \bottomrule
    \end{tabular}}
    \caption{The impact of delta pose sequence length $L_{d}$, sequence step $s$, and delta step $s_{d}$.}
    \label{table:Ablation_length_step}
    \end{subtable}

    \caption{Ablation Studies on the variations of pose condition inputs. The result is on I3D-Human.}
    \label{table:Ablation}
\end{table}

\section{Ablation Studies}
\subsection{Pose Sequence Condition}
We ablate the design of the dynamic context sequence input, including the following factors.

\smallskip \noindent \textbf{Pose Sequence or Delta Pose Sequence:} As for each element in the input context sequence, instead of the pose parameter at each step, we use the pose difference between two consecutive steps including the angular and translational velocities. We call the resulting sequence as the delta pose sequence $S_{\Delta{\bm{p}}}$. We compare inputting pose sequence $S_{\bm{p}}$ and delta pose sequence $S_{\Delta{\bm{p}}}$, and show the results on I3D-Human in~\cref{table:Ablation_delta}. Inputting $S_{\Delta{\bm{p}}}$ achieves better performance on both novel view and novel pose, thanks to the reduction of the input complexity and comprehensive inclusion of all dynamic contexts.

\smallskip \noindent \textbf{Delta Pose Representation:} In our model, we represent the angular velocities of the joints by axis-angles. We also experiment with other delta pose representations and compare them in~\cref{table:Ablation_rot-rep}. Due to its low-dimensional representation and lack of ambiguity, the axis-angle representation outperform the Rodrigues and quaternion representations on both novel view and novel pose. 

\smallskip \noindent \textbf{Length and Step:} We further ablate on different choices of delta pose sequence length $L_{d}$, sequence step $s$, and delta step $s_{d}$. We set $s$ and $s_{d}$ the same in our experiments. The ablation results can be found in~\cref{table:Ablation_length_step}. Short sequence ($L_{d}=2$) lacks sufficient dynamic contexts, leading to a decrease in performance on both novel view and novel pose. On the other hand, long sequence ($L_{d}=12$) contains overly complex input, which leads to comparable performance on novel view but worse overfitting performance on novel pose. Reducing $s$ and $s_{d}$ ($s \,\& \,s_d=12$) similarly leads to a loss of dynamic contexts, resulting in deteriorated performance on both novel view and novel pose. We rigorously select the most suitable parameter combinations to encompass an adequate amount of dynamic contexts while mitigating overfitting induced by complex inputs.

\subsection{The effectiveness of Localized Spatial Dependency}
\label{sec:local-dependency}
\begin{figure}[t!]
    \centering
    \includegraphics[width=0.9\linewidth]{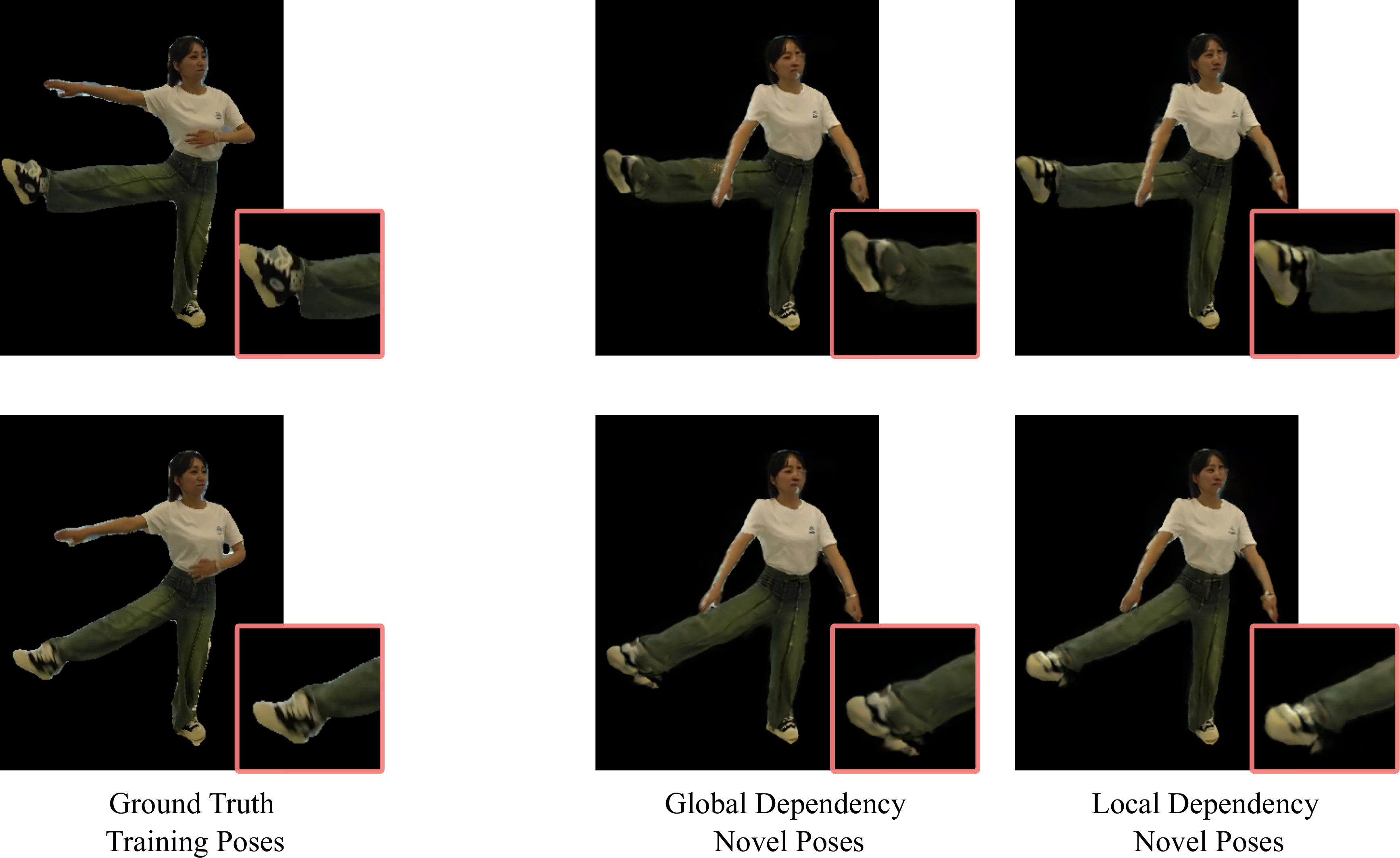}
    \caption{Localized spatial dependency enhances generalization for novel poses composed of observed sub-poses. Consider a training pose (left) from which we create a novel pose (right) by substituting the upper limbs' pose with that from another training frame. This novel combination of sub-poses challenges models with global dependency, leading to inaccuracies in rendering dynamic-related details on the pant legs. In contrast, our method with local pose dependency can still capture the appearance variation.}
    \label{fig:cmp_novelpose_glocal}
\end{figure}
 We quantitatively measure the impact of local or global spatial dependency in Tab. 3 of the main text, where the last two rows indicate that localization improves three conventional metrics on the test set. However, we observe that localization does not enhance DME. This may be due to: (1) DME computation relying on optical flow estimation, which may contain errors, and (2) our dataset's lack of diverse pose combinations, which limits the ability to highlight the importance of avoiding spurious inter-body dependency. To address this limitation and prove the effectiveness of localized spatial dependency, we examine novel pose combinations by recombining sub-poses observed in the training set. As illustrated in~\cref{fig:cmp_novelpose_glocal}, suppose we have a video sequence where the character waves her arms and legs in opposite directions. We create a novel pose by substituting the sub-pose of the upper limbs with the one from another time-step. Thanks to the observed sub-poses and the limited interdependence between the upper and lower limbs, it is expected that they should be rendered akin to their appearance in the training frames under this novel pose combination. We compare renderings by two models with local and global spatial dependency. We show that the one with local dependency effectively captures the local motion of the pant leg, while the other one with global dependency fails to accurately render the legs under the novel pose combination as the encoded global context condition is unseen.

\begin{figure}[t]
    \centering
    \includegraphics[width=.9\textwidth]{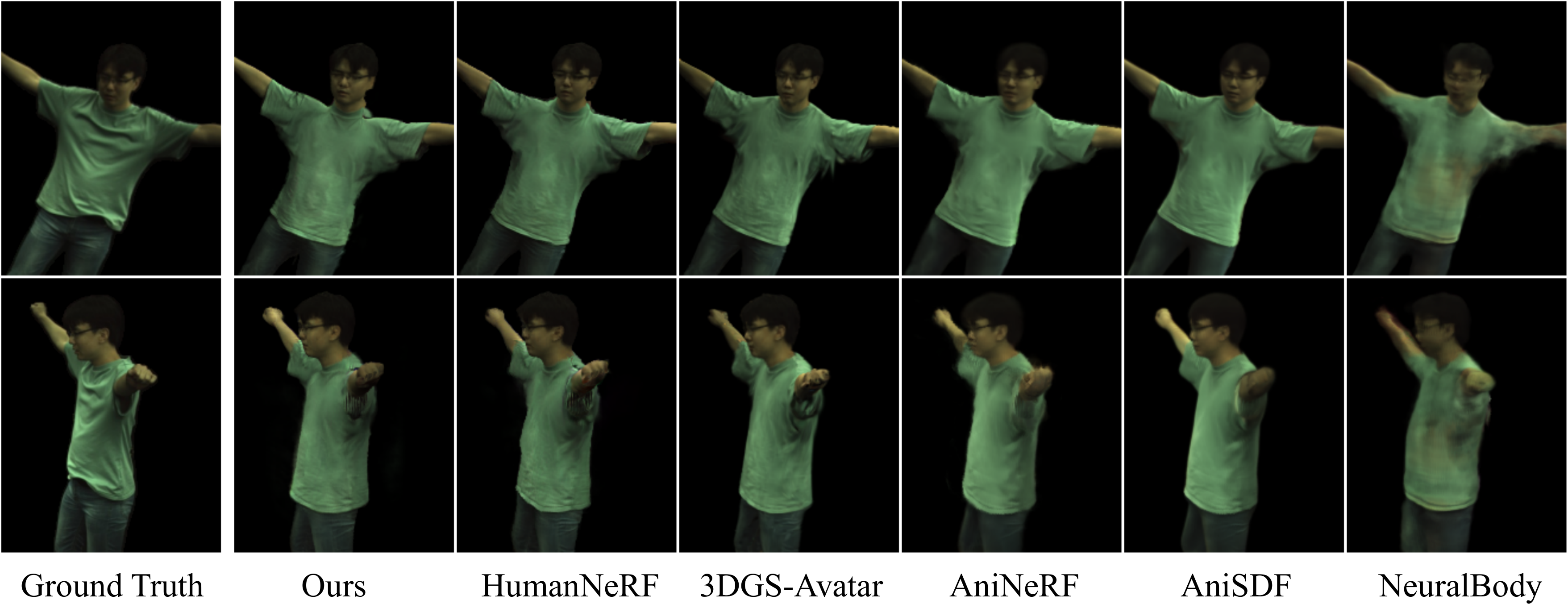}
    \caption{Existing implicit avatars have difficulties in rendering out-of-distribution poses in the ZJU-MoCap dataset. When the character lifts the limbs, all models fail to accurately predict the deformation and appearance of the underarm as it was not observed during training.}
    \label{fig:zju_novelpose-failure}
\end{figure}

\begin{figure}[h]
    \centering
    \includegraphics[width=1.\textwidth]{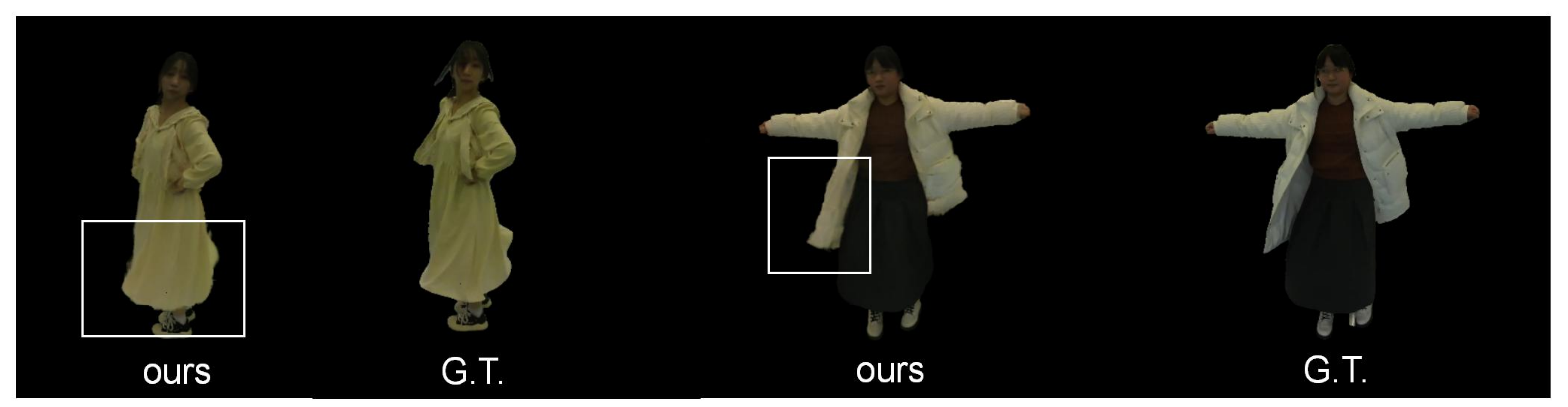}
    \caption{Our failure mode in I3D-Human. The performance degrades when extrapolating to out-of-distribution dynamics, such as significant motion with high speed.}
    \label{fig:failure case in I3D-Human}
\end{figure}

\section{Limitations}
Our method (Dyco) is the first implicit avatar model that enables the rendering of diverse dynamic motion effects under similar static poses with different motion histories. This capability is crucial for enhancing the rendering fidelity of loosely-clothed moving avatars. However, Dyco also has its limitations.

Firstly, similar to previous implicit avatars, Dyco encounters challenges when animating out-of-distribution novel poses.  Although we propose localized spatial dependency to help generalization towards unseen poses composed of seen sub-poses, as shown in \cref{fig:cmp_novelpose_glocal}, it still struggles in predicting appearance and geometry variation under completely novel poses and inpainting unseen body parts. We provide an illustrative example from the ZJU-MoCap novel pose test set in~\cref{fig:zju_novelpose-failure}, where the character lifts his elbow, revealing the underarm region that was not observed during training. It's important to note that other methods also fail to render the underarm plausibly. 

Secondly, although our method is able to produce plausible dynamic effects within a range of motion velocities, the performance degrades when extrapolating to out-of-distribution dynamics, such as the rising of a drape during extremely high-speed spinning, as shown in~\cref{fig:failure case in I3D-Human}. Acquiring some training data encompassing the extreme case may help enhance the generalization. Additionally, integrating explicit physical rules to model the relationship between speed and motion offers another potential solution.

Thirdly, our current method does not consider random environmental factors, such as wind blowing. We posit that the multi-modal appearance under similar static poses primarily arises from the motion context of human kinematics. External forces, like wind, could be modeled as random noise or through specially designed physical mechanisms. We leave the exploration of this aspect for future work.

Lastly, the proposed dynamic motion error (DME) metric relies on the accurate estimation of 2D optical flow. However, we find that off-the-shelf optical flow estimators, such as RAFT~\cite{Teed2020_RAFT}, are not perfect. While DME can generally measure how accurately the rendering exhibits motion dynamics, as shown in~\cref{fig:dme_cmp}, it may not strictly correlate with performance at a finer scale, as shown in some of our ablation results. Research progress in 3D point tracking might shed light on the measurement of motion dynamics.

\begin{figure}[ht]
    \centering
    \begin{subfigure}{\textwidth}
     \includegraphics[width=0.95\textwidth]{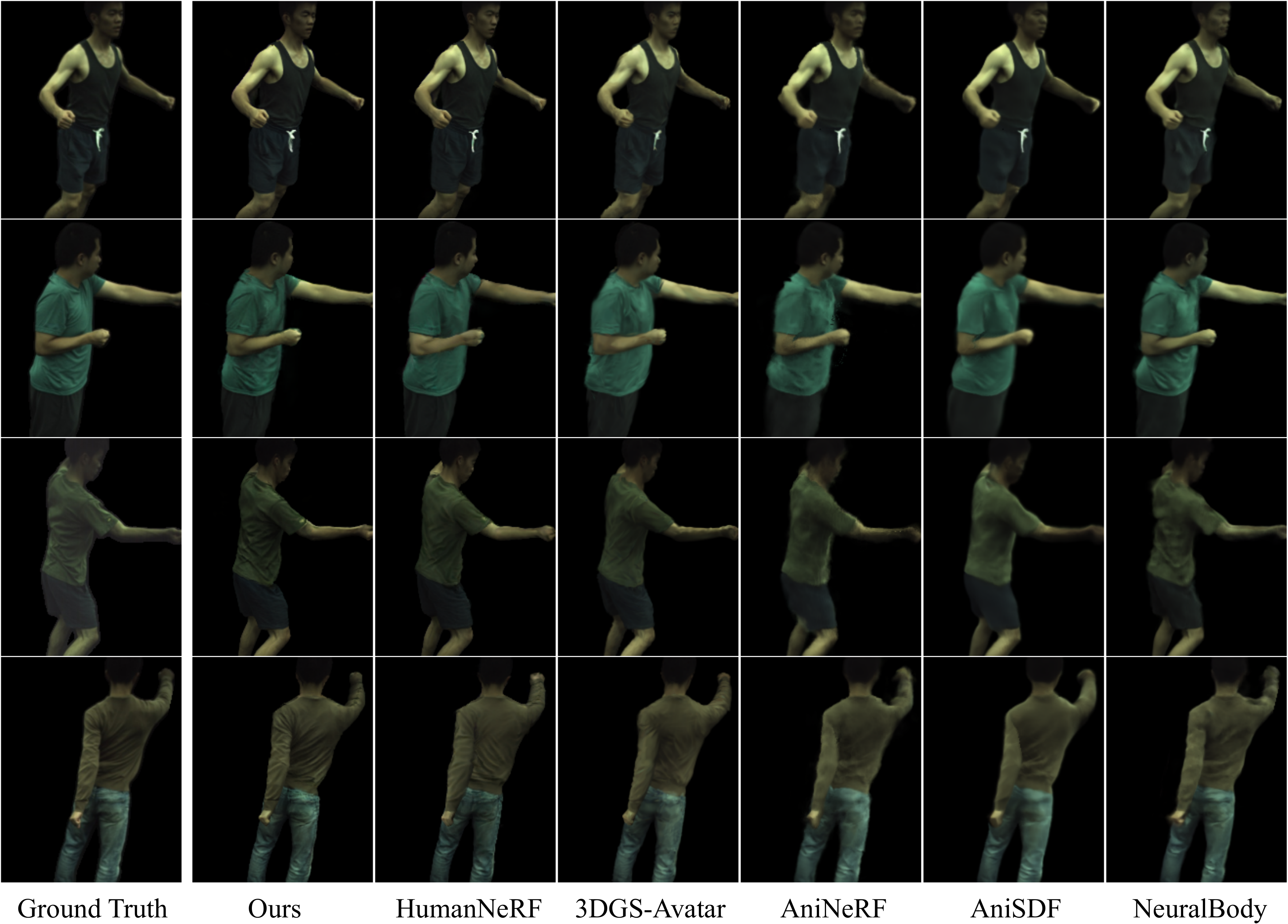}
    \caption{Novel View}
    \label{fig:zju_novelview}
    \end{subfigure}
    \begin{subfigure}{\textwidth}
     \includegraphics[width=0.95\textwidth]{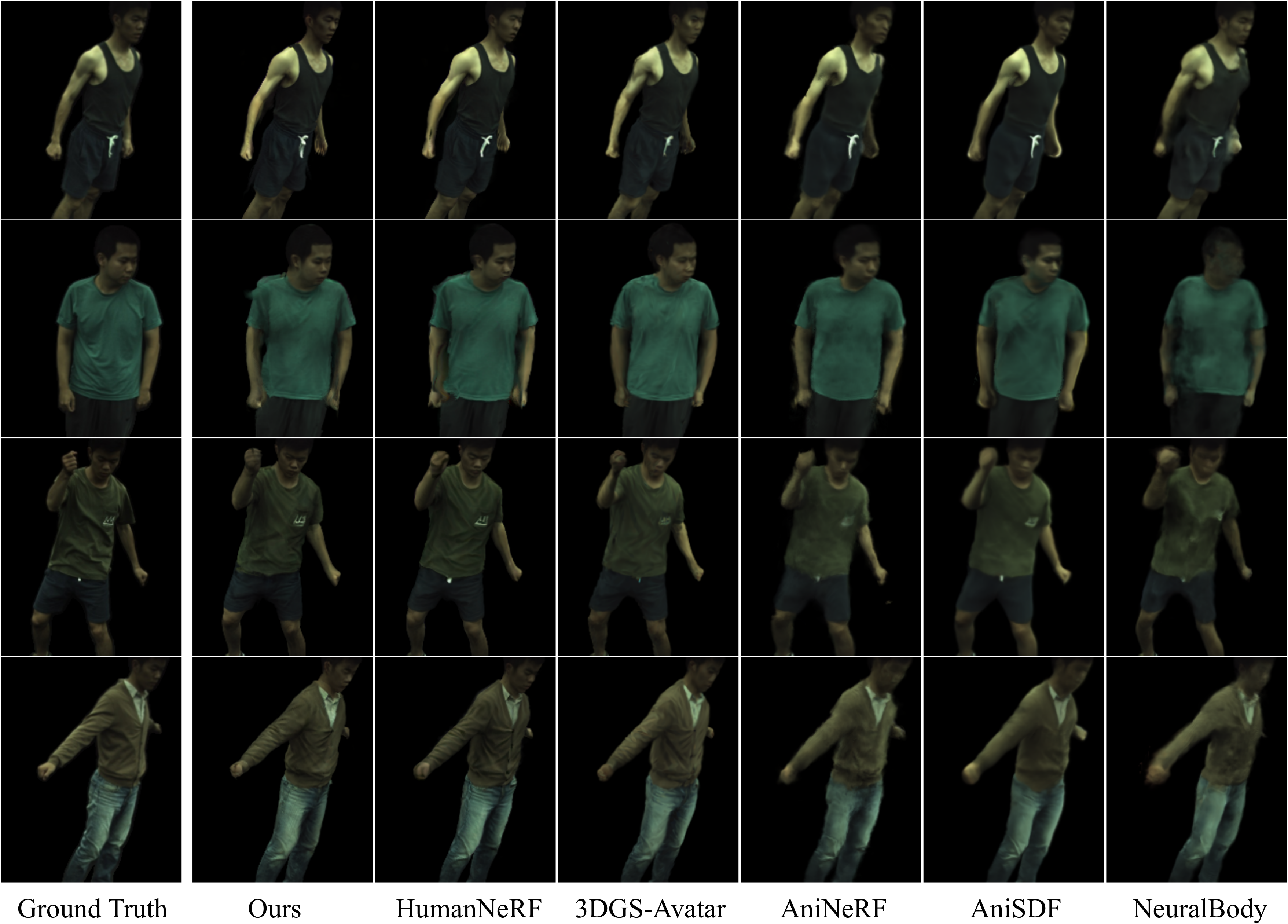}
    \caption{Novel Pose}
    \label{fig:zju_novelpose}
    \end{subfigure}
    \caption{Qualitative comparison on the ZJU-MoCap dataset. Our method captures more details than others in the novel-view test set and performs comparably with others in the novel-pose test set. }
    \label{fig:zju_cmp}
\end{figure}


\end{document}